\documentclass[letterpaper, 10 pt, conference]{ieeeconf}
\IEEEoverridecommandlockouts 
\usepackage{hyperref, url, amsmath, amssymb, booktabs, graphicx, bm, color, algorithm, tabularx}
\usepackage[noend]{algpseudocode}
\usepackage[caption=false]{subfig}
\graphicspath{{images/}}

\newcolumntype{Y}{>{\centering\arraybackslash}X}

\title{\LARGE \bf Multidimensional Capacitive Sensing \\ for Robot-Assisted Dressing and Bathing}

\author{Zackory Erickson, Henry M. Clever, Vamsee Gangaram, Greg Turk, C. Karen Liu, and Charles C. Kemp
\thanks{Zackory Erickson, Henry M. Clever, Vamsee Gangaram, and Charles C. Kemp are with the Healthcare Robotics Lab, Georgia Institute of Technology, Atlanta, GA., USA.}%
\thanks{C. Karen Liu and Greg Turk are with the School of Interactive Computing, Georgia Institute of Technology, Atlanta, GA., USA.}%
\thanks{Zackory Erickson is the corresponding author {\tt\footnotesize zackory@gatech.edu}.}%
}

\begin{document}

\maketitle
\thispagestyle{empty}
\pagestyle{empty}

\begin{abstract}
Robotic assistance presents an opportunity to benefit the lives of many people with physical disabilities, yet accurately sensing the human body and tracking human motion remain difficult for robots. We present a multidimensional capacitive sensing technique that estimates the local pose of a human limb in real time. A key benefit of this sensing method is that it can sense the limb through opaque materials, including fabrics and wet cloth. Our method uses a multielectrode capacitive sensor mounted to a robot's end effector. A neural network model estimates the position of the closest point on a person's limb and the orientation of the limb's central axis relative to the sensor's frame of reference. These pose estimates enable the robot to move its end effector with respect to the limb using feedback control. We demonstrate that a PR2 robot can use this approach with a custom six electrode capacitive sensor to assist with two activities of daily living---dressing and bathing. The robot pulled the sleeve of a hospital gown onto able-bodied participants' right arms, while tracking human motion. When assisting with bathing, the robot moved a soft wet washcloth to follow the contours of able-bodied participants' limbs, cleaning their surfaces. Overall, we found that multidimensional capacitive sensing presents a promising approach for robots to sense and track the human body during assistive tasks that require physical human-robot interaction.
\end{abstract}

\section{Introduction}
\label{sec:intro}


Robots that provide direct physical assistance offer an opportunity to positively impact the lives of many people who require support with everyday tasks.
For example, the US Census Bureau has projected that by 2030, over 20\% of the US population will be over the age of 65~\cite{vespa2018demographic}.
Of older adults over 65, $\sim$40\% report having some form of disability~\cite{mitzner2014identifying}. People receiving assistance at home from formal caregivers most frequently report receiving assistance with dressing and bathing compared to other activities of daily living (ADLs) and instrumental activities of daily living (IADLs)~\cite{mitzner2014identifying}. Yet, physical human-robot interaction related to these tasks presents several challenges for robots, in part due to difficulties in sensing the human body. Enabling a robot to better estimate human pose and track body motion could be advantageous for many scenarios in which robots and humans physically interact.


In this paper, we present a technique that uses multidimensional capacitive sensing to estimate the relative pose of a person's limb. Our implementation uses a capacitive sensor mounted on the end effector of a mobile manipulator that consists of 6 conductive electrodes in a 3 $\times$ 2 array.
We use a data-driven model that, when given a series of measurements from this capacitive sensor, estimates the relative position of the closest point on the surface of a person's limb, $\bm{p} = (p^{}_y, p^{}_z)$, as well as the pitch and yaw orientation, $\bm{\theta} = (\theta^{}_y, \theta^{}_z)$, between a robot's end effector and the central axis of the limb near the point.

When combined with feedback control, we show that this sensing approach enables a robot to track human movement and follow the contours of a person's body while providing assistance with tasks such as dressing and bathing.
In addition, our approach provides high sampling rates, with low latency, which allows the robot to estimate a person's local pose at over 100 Hz.
Unlike computer vision approaches for estimating a person's pose, our capacitive sensing technique can sense the human body through some materials relevant to assistance, including fabric and wet cloth, which can inhibit a robot's visual sight of a person during assistive tasks, such as dressing~\cite{gao2015user}. Furthermore, while current robotic systems have used capacitive sensing at a robot's end effector to follow the 3D contours of static objects~\cite{navarro20146d}, no prior works have explored how a robot can use capacitive sensing to track the position and orientation of a dynamic human limb while providing physical assistance.

\begin{figure}
\centering
\includegraphics[width=0.48\textwidth, trim={30cm 13cm 18cm 0cm}, clip]{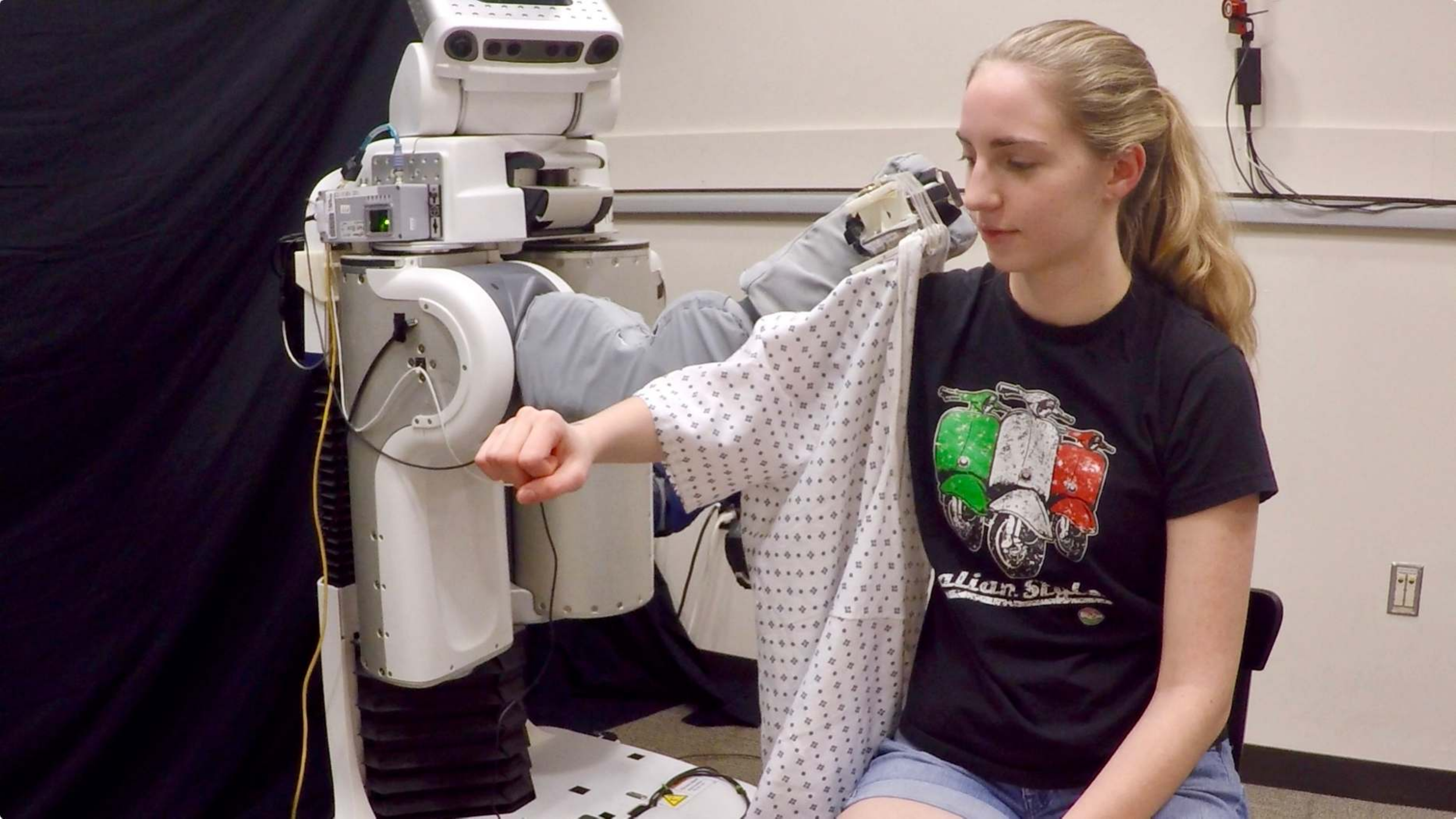}
\vspace{-0.4cm}
\caption{\label{fig:intro}A PR2 robot uses multidimensional capacitive sensing to track human motion while pulling a hospital gown onto a participant's arm.}
\vspace{-0.4cm}
\end{figure}

\begin{figure}
\centering
\includegraphics[width=0.48\textwidth, trim={0cm 0cm 0cm 10cm}, clip]{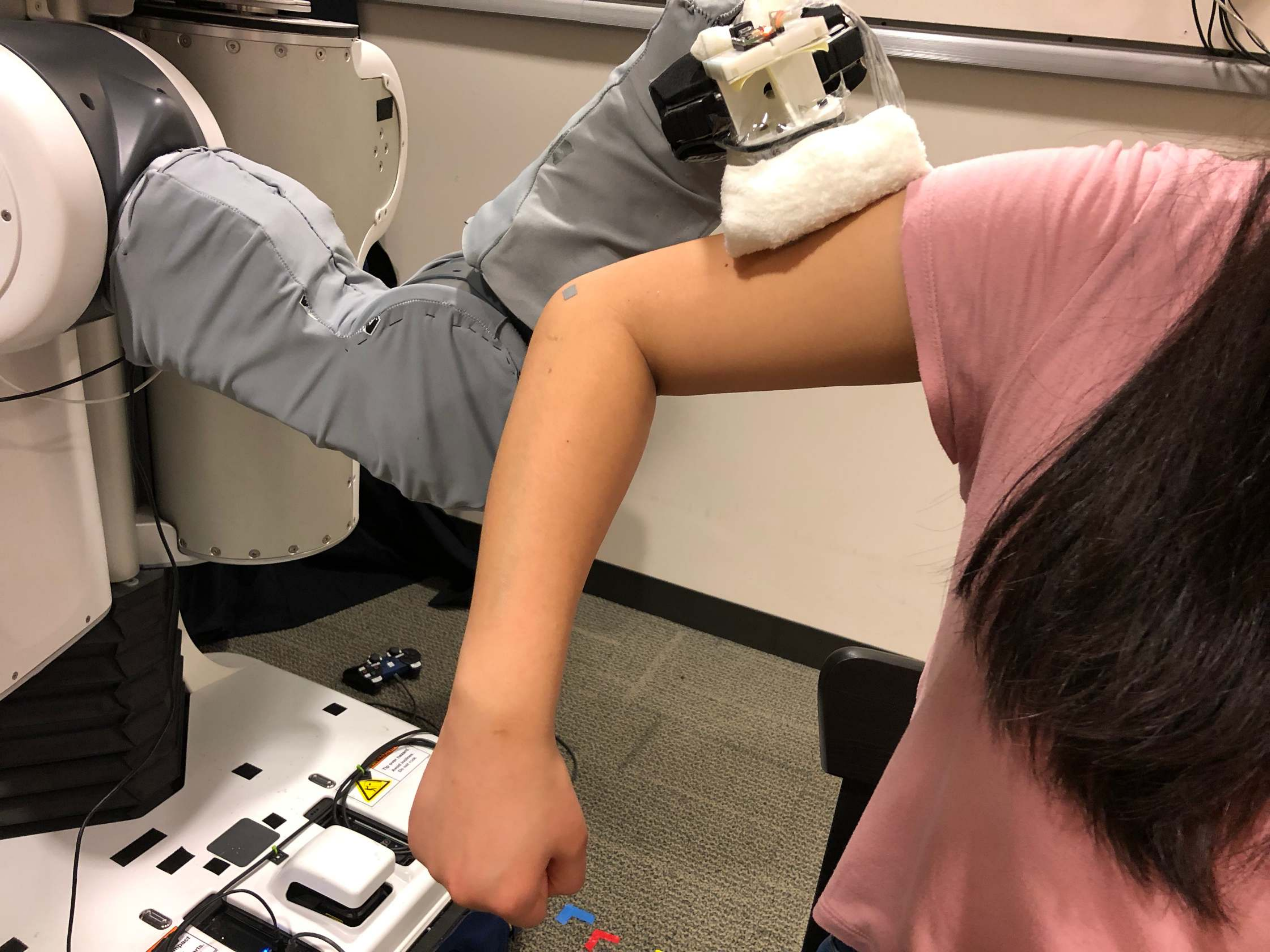}
\vspace{-0.4cm}
\caption{\label{fig:intro_wiping}A PR2 robot uses our capacitive sensing approach with a wet washcloth to follow and clean a participant's arm.}
\vspace{-0.4cm}
\end{figure}

We demonstrate our approach in a study with able-bodied human participants during which a robot used our capacitive sensing technique to assist in fully dressing the sleeve of a hospital gown onto participants' arms, as depicted in Fig.~\ref{fig:intro}. We show how our method enables a robot to track human motion, including translation and rotation of a human limb, which allowed our robot to successfully dress participants even in the presence of unscripted human arm motion. Furthermore, we demonstrate how this sensing approach can generalize to assisting with another task---bathing---during which the robot used a wet cloth to clean the outer surfaces of participants' arms and legs, as seen in Fig.~\ref{fig:intro_wiping}.


Through this work, we make the following contributions:
\begin{itemize}
\item We propose and evaluate a capacitive sensor design capable of sensing the local pose of a human limb.
\item We present a data-driven model trained on capacitance measurements from a single user and we show that this model can generalize across multiple people to estimate the position of the closest point on a person's limb and the orientation of the limb's central axis.
\item When combined with feedback control, we demonstrate how a robot can leverage capacitive sensing to track human motion and assist with two real-world tasks: dressing the sleeve of a hospital gown, and cleaning the arm and leg with a wet washcloth.
\end{itemize}

\section{Related Work}
\label{sec:related_work}

\subsection{Capacitive Sensing}

Within robotics, one prominent use of capacitive sensing is for proximity sensing of objects and the human body~\cite{kirchner2008capacitive, goeger2010tactile}. Navarro et al. presented a two parallel plate design with 2~$\times$~2 capacitive sensor arrays that could follow the curvature and orientation of stationary objects that fit in between the two plates~\cite{navarro20146d}.
However, this approach does not extend to sensing human limbs larger than the gap between two fixed parallel plate sensors.
In comparison, our approach uses a single-sided capacitive sensor array to track dynamic human movement across human limbs of varying shapes and sizes.

Researchers have also looked at capacitive sensing for improving human-robot interaction. Lee et al. designed a 16~$\times$~16 array of capacitive sensors which could be used as a robotic skin to sense the presence of a human hand from 17~cm away~\cite{lee2009dual}. Hoffmann et al. mounted capacitive sensors to the surface of two industrial robotic arms, which the robots could use to detect and avoid people in the workspace~\cite{hoffmann2016environment}. Additionally, Navarro et al. presented a capacitive sensor design for safer human-robot interaction, consisting of a table-mounted 3 by 16 grid of sensors for estimating the location and proximity of a nearby human hand~\cite{navarro2013methods}. 


In prior work, we introduced a capacitive sensor for sensing one-dimensional vertical distance between a robot's end effector and a human arm, which enabled a robot to track vertical arm motion while dressing the sleeves of both a hospital gown and a  long-sleeve sweater~\cite{erickson2018tracking}. This prior approach used a model inspired by the capacitance equation for a parallel plate capacitor, $d(C) = \frac{\alpha}{C + \beta}$, which estimated the distance between a sensor and a human limb given a single capacitance measurement, $C$, and constants, $(\alpha, \beta)$, obtained through a least-squares fit.
However, this model lacks a clear generalization to estimating the pose of a person's limb with multiple capacitive sensor inputs, due in part to crosstalk among multiple electrodes in close proximity to one another.
Furthermore, no previous work has demonstrated how a robot can use capacitive sensing to track 4-dimensional human limb pose during physical human-robot interaction.

\begin{figure*}
\centering
\subfloat[\label{subfig:a}]{\includegraphics[width=0.23\textwidth, trim={11cm 4.55cm 9cm 6cm}, clip]{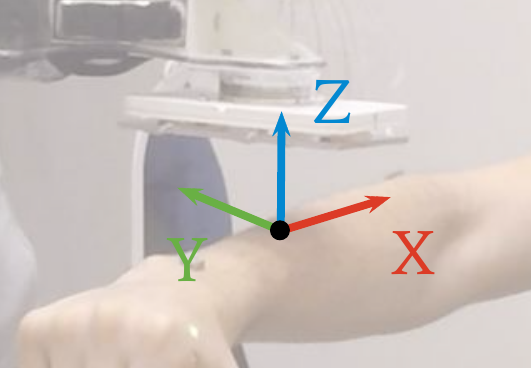}}\ 
\subfloat[\label{subfig:b}]{\includegraphics[width=0.23\textwidth, trim={6cm 3.3cm 5cm 1cm}, clip]{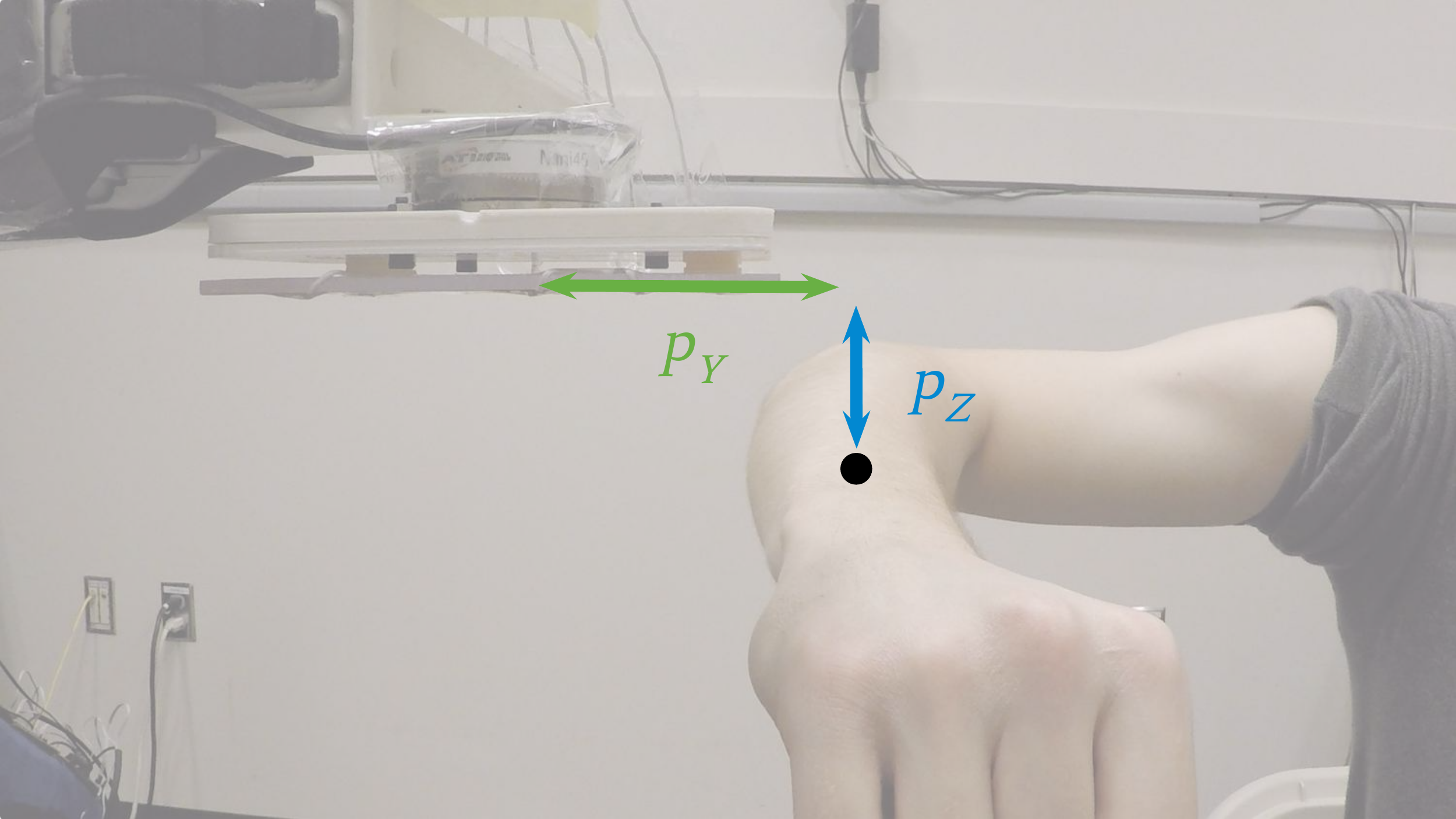}}\ 
\subfloat[\label{subfig:c}]{\includegraphics[width=0.23\textwidth, trim={6cm 3cm 6cm 2cm}, clip]{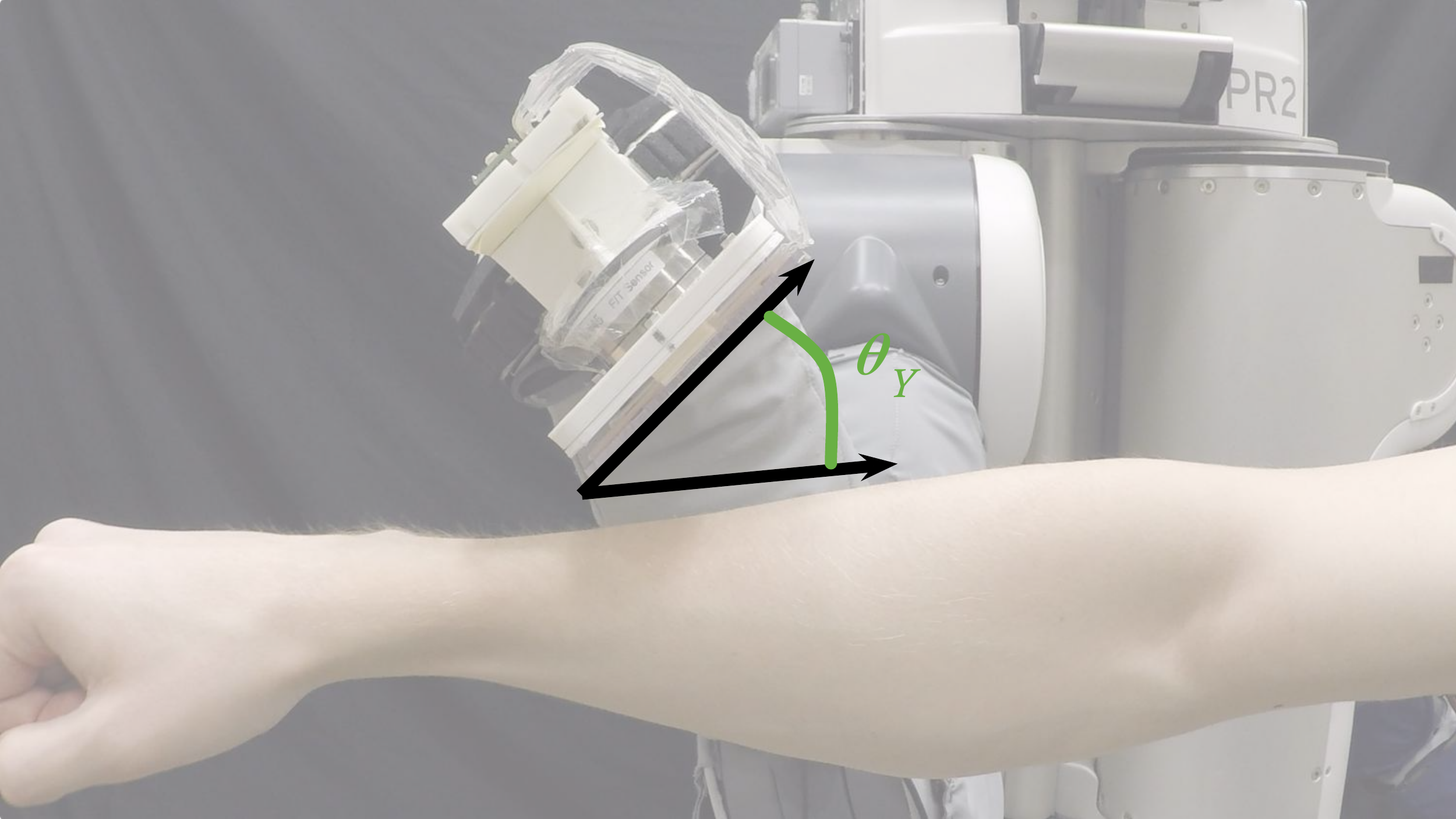}}\ 
\subfloat[\label{subfig:d}]{\includegraphics[width=0.23\textwidth, trim={3cm 1cm 9cm 4cm}, clip]{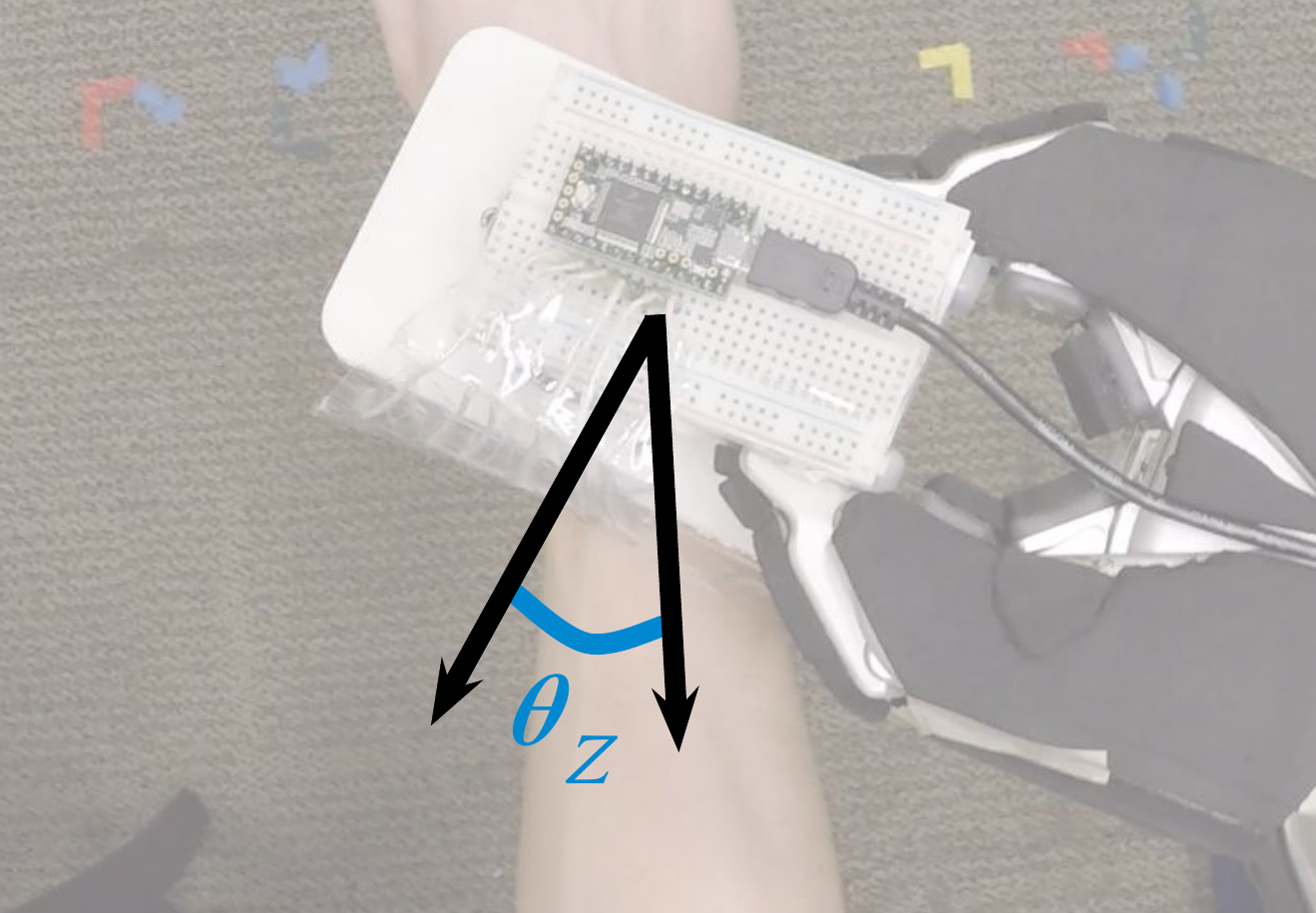}}\ 
\caption{\label{fig:axes}(a) The local coordinate frame on the limb at the closest point to the end effector. (b) The lateral and vertical position differences between the end effector and the closest point on the limb. (c) Pitch orientation difference between the end effector and the central axis of the limb. (d) Yaw orientation to the central axis of the limb.}
\vspace{-0.4cm}
\end{figure*}

\begin{figure}
\centering
\includegraphics[width=0.23\textwidth, trim={15cm 18.4cm 10cm 3cm}, clip]{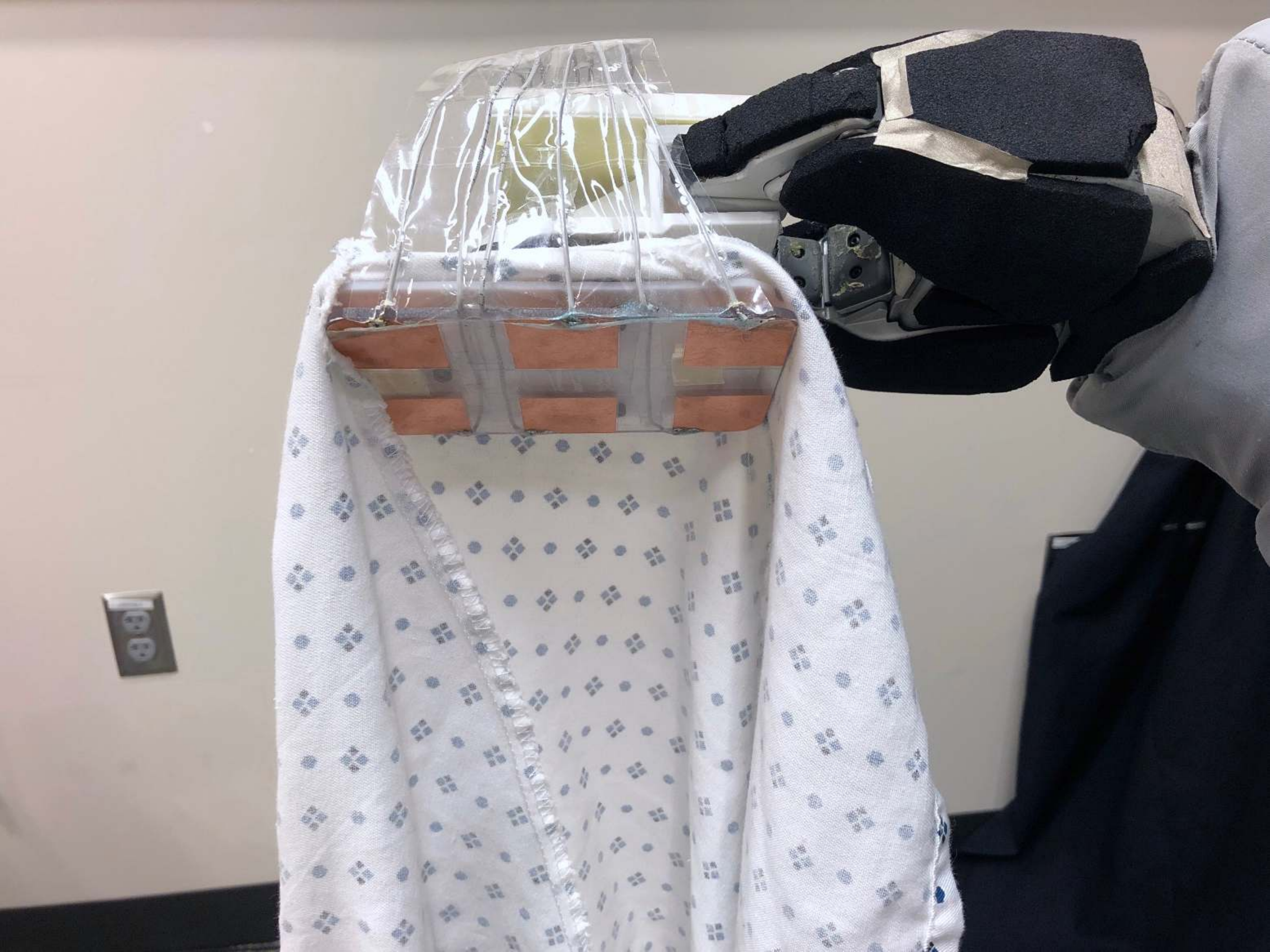}
\includegraphics[width=0.23\textwidth, trim={15cm 34.5cm 30cm 3cm}, clip]{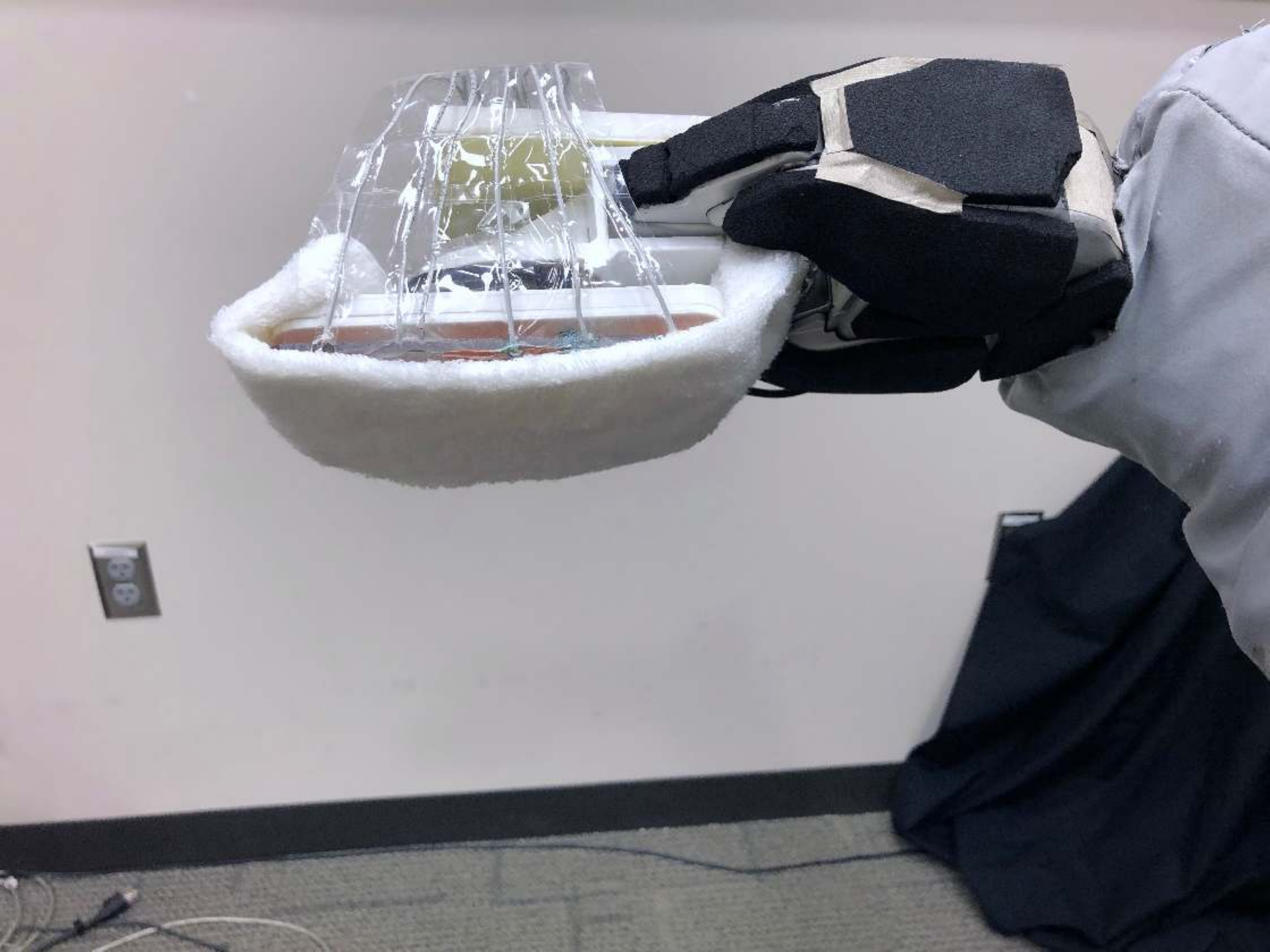}\\
\vspace{0.25em}
\includegraphics[width=0.23\textwidth, trim={20cm 15cm 15cm 15cm}, clip]{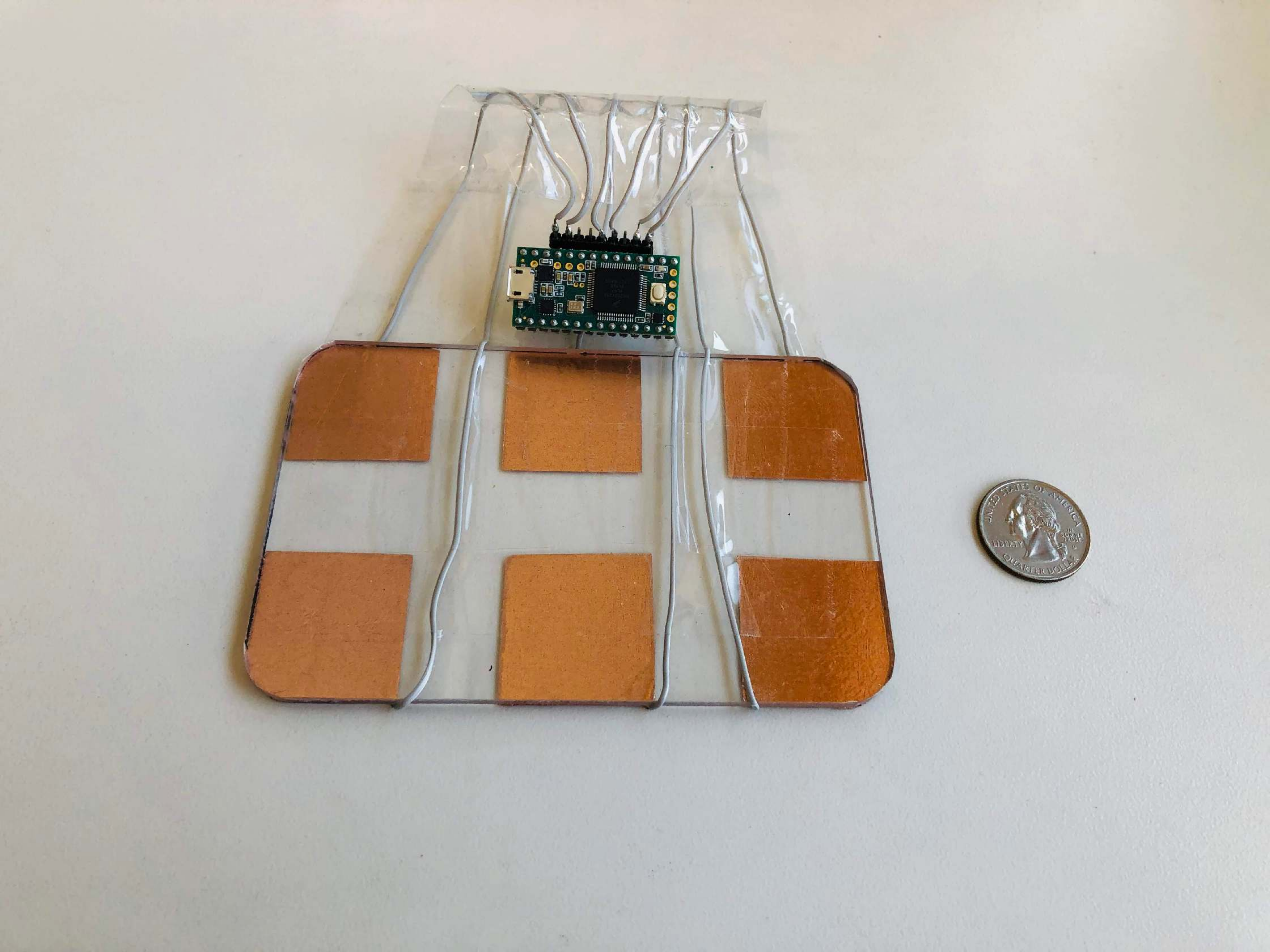}
\caption{\label{fig:sensor}Bottom: Six electrode capacitive sensor connected to the Teensy microcontroller board. The capacitive sensor is then mounted onto the bottom of a tool that the PR2 holds. Top left: The tool holding a hospital gown. Top right: The tool with a wet washcloth attached under the sensors.}
\vspace{-0.4cm}
\end{figure}

\subsection{Robot-Assisted Dressing and Bed Bathing}

Research on recovering visually occluded human pose in healthcare environments has employed a variety of sensing techniques, including force sensors~\cite{gao2016iterative}, pressure mats~\cite{clever20183d}, and feature extraction from non-occluded joints in partially occluded images~\cite{chance2018elbows}.
In comparison to visual feedback approaches, capacitive sensing enables a robot to sense the human body through some opaque materials, such as clothing~\cite{erickson2018tracking}.
When compared to force feedback, capacitive sensing can track human motion \textit{before} the robot applies forces to a person's body.

Many approaches have used vision systems, such as the Microsoft Kinect, to track human pose for robot-assisted dressing~\cite{pignat2017learning}. 
Koganti et al. used the Kinect depth camera and motion capture to estimate the cloth state of a T-shirt as a Baxter robot pulled the T-shirt over a mannequin's head and torso~\cite{koganti2017bayesian}. 
Chance et al. also used a depth camera with a recurrent neural network to estimate the pose of an occluded human arm as a Baxter assisted in pulling on a jacket~\cite{chance2018elbows}.

Recent work has also used force sensing to improve robot-assisted dressing performance and reduce the forces applied onto a person's body~\cite{zhang2017personalized}. Gao et al. proposed a stochastic path optimization approach to adjust a Baxter's motion given force feedback, while dressing a sleeveless jacket~\cite{gao2016iterative}. 
Chance et al. used a 6-axis force/torque sensor to identify dressing errors and clothing types when dressing human participants~\cite{chance2017quantitative}. In prior work, we presented a deep haptic model predictive control approach for robot-assisted dressing, that enabled a robot to predict the future forces a garment would apply to a person's body given force/torque feedback from a robot's end effector~\cite{erickson2018deep}.


Relatively few researchers have explored robotic assistance for bathing. Satoh et al.~\cite{satoh2009bathing} demonstrated how a robotic exoskeleton could help with carrying a person to a bathing location, and Bezarra et al.~\cite{bezarra2017bath} developed a mechatronic system to control bath water. One common bathing strategy within nursing care are bag baths, which are often performed in bed and do not require large amounts of water or drying~\cite{carruth1995bag}. Zlatintsi et al. presented a multimodal learning framework for recognizing a person's audio and gesture commands to improve interaction with a robotic bathing system~\cite{zlatintsi2018multimodal}. King et al. presented a robotic bed bathing system that wiped small patches of debris off of a person's static arm and leg~\cite{king2010towards}. Unlike our approach, the authors used a compliant force-controlled wiping behavior to clean user-selected areas on a person's body and did not estimate human pose or track human motion.

\section{Sensor, Model, and Training}
\label{sec:method}

In this section, we present the multidimensional capacitive sensor design that enables a robot to follow the surface contours of the human body (Fig.~\ref{fig:axes}) and track human limb movement. In addition, we introduce our data collection process, our model used for estimation, and our robot controller used throughout our experiments with human participants.

\subsection{Capacitive Sensor Design}
\label{sec:sensor}

Our capacitive sensor design consists of six positively charged capacitive electrodes, which were adhered to an assistive tool held by a robot's end effector. Fig.~\ref{fig:sensor} displays the capacitive sensor by itself, as well as affixed to the bottom of the tool with a fabric hospital gown or wet washcloth. Each of the sensor electrodes is a 3~cm $\times$ 3~cm square of copper foil mounted in a 3 by 2 grid arrangement. This tool also contains an ATI 6-axis force/torque sensor, which allowed the robot to monitor the force applied to a person's body in the event of direct contact. 

We connected the six capacitive electrodes to a Teensy 3.2 microcontroller board, a commercially available board with built-in components for capacitive proximity sensing, including a 1~pF internal reference capacitor. The Teensy is capable of sampling capacitance measurements from all six electrodes at frequencies over 100 Hz.



\subsection{Estimating Position/Orientation of a Human Limb}

\begin{figure}
\centering
\includegraphics[width=0.48\textwidth, trim={0.15cm 0cm 0.15cm 0cm}, clip]{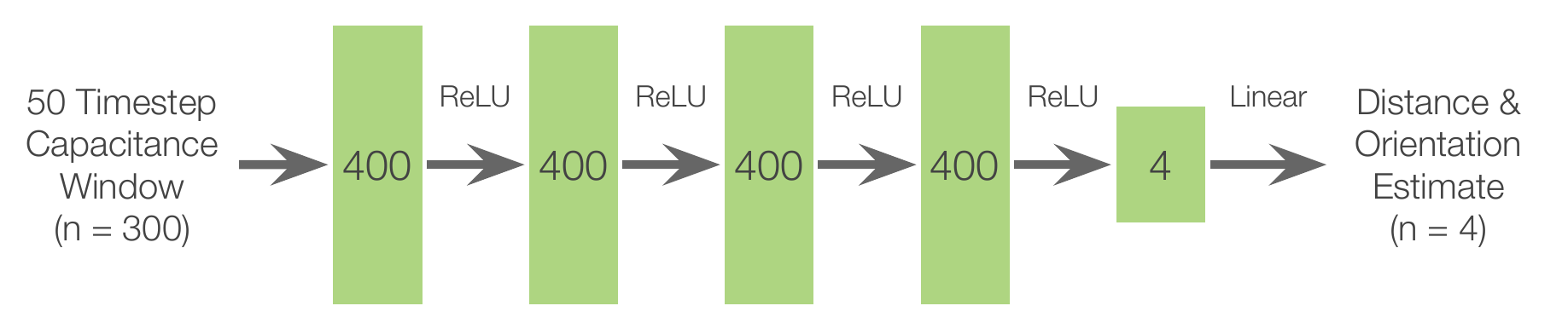}
\vspace{-0.6cm}
\caption{\label{fig:nn}Neural network architecture. We apply a ReLU activation after each hidden layer.}
\vspace{-0.4cm}
\end{figure}

Given capacitance measurements from this sensor, we aim to inform a robot about the local pose of a person's limb.

We design a model, $f(\bm{c}_{t-h+1:t})$, that takes as input a window of prior capacitance measurements, $\bm{c}^{}_{t-h+1:t}$, from time step $t-h+1$ to the current time $t$. Our model then outputs, $\bm{y}^{}_t = (p^{}_{t,y}, p^{}_{t,z}, \theta^{}_{t,y}, \theta^{}_{t,z})$, which includes estimates of the relative position of the closest point on the surface of a person's limb, $\bm{p}^{}_t = (p^{}_{t,y}, p^{}_{t,z})$, and 
orientation of the limb's central axis relative to the sensor's frame of reference, $\bm{\theta}^{}_t = (\theta^{}_{t,y}, \theta^{}_{t,z})$, as shown in Fig.~\ref{fig:axes}. We do not estimate the relative position or orientation with respect to the X-axis of a participant's limb, as these two degrees of freedom are difficult to measure for a limb that is approximately cylindrical in shape.

In this work, we use a time window over the last 50 time steps ($h=50$), or 0.5 seconds given a sampling frequency of 100 Hz. With six capacitance measurements per time step, our input $\bm{c}^{}_{t-49:t}\in\mathbb{R}^{50\times 6}$ can be vectorized as a 300-dimensional vector, i.e. $f: \mathbb{R}^{300} \rightarrow \mathbb{R}^4$.

To estimate position and orientation, we trained a fully-connected neural network model, with the architecture shown in Fig.~\ref{fig:nn}.
Our model takes as input a 300-dimensional vector of capacitance measurements, $\bm{c}^{}_{t-49:t}$. Our model, trained in Keras with TensorFlow, consists of four 400 node layers with ReLU activations and a final four node layer which outputs the estimated $\bm{\hat{y}}^{}_t$. We trained the model for 100 epochs with a batch size of 128. We used the Adam optimizer with $\beta_1=$~0.9, $\beta_2=$~0.999, and a learning rate of 0.001.

\subsection{Training a Model}
\label{sec:generalized_model}

\begin{algorithm}[t]
\caption{Data Collection}\label{alg:datacollect}
\begin{algorithmic}[1]
\State \textbf{input:} time window $h$, $N$ iterations.
\State $\mathcal{D} \gets \{\}$
\For{$i=1,\ldots, N$}
\State Select target $\bm{y}^{}_T = (p^{}_{T,y}, p^{}_{T,z}, \theta^{}_{T,y}, \theta^{}_{T,z})\in S$.
\State Select velocities $\bm{v}^{}_p = (v^{}_{p_y}, v^{}_{p_z})$ s.t. $||\bm{v}^{}_p|| \in [3, 10]$.
\State Select velocities $\bm{v}^{}_\theta = (v^{}_{\theta_y}, v^{}_{\theta_z})$ s.t. $||\bm{v}^{}_\theta|| \in [\frac{\pi}{20}, \frac{\pi}{8}]$.
\State $t \gets 0$.
\While{$\bm{y}^{}_t \not\approx \bm{y}^{}_T$}
\State Observe capacitance measurements $\bm{c}^{}_t \in \mathbb{R}^6$.
\State Observe $\bm{y}^{}_t$ using forward kinematics.
\If{$t \geq h$}
\State $\mathcal{D} \gets \mathcal{D} \cup \{(\bm{c}^{}_{t-h+1:t}, \bm{y}^{}_t)\}$
\EndIf
\State Take action towards $y^{}_T$ with velocities $\bm{v}^{}_p, \bm{v}^{}_\theta$.
\State $t \gets t+1$.
\EndWhile
\EndFor
\State \Return dataset $\mathcal{D}$
\end{algorithmic}
\end{algorithm}

\begin{figure}
\centering
\includegraphics[width=0.23\textwidth, trim={5cm 3cm 7cm 0cm}, clip]{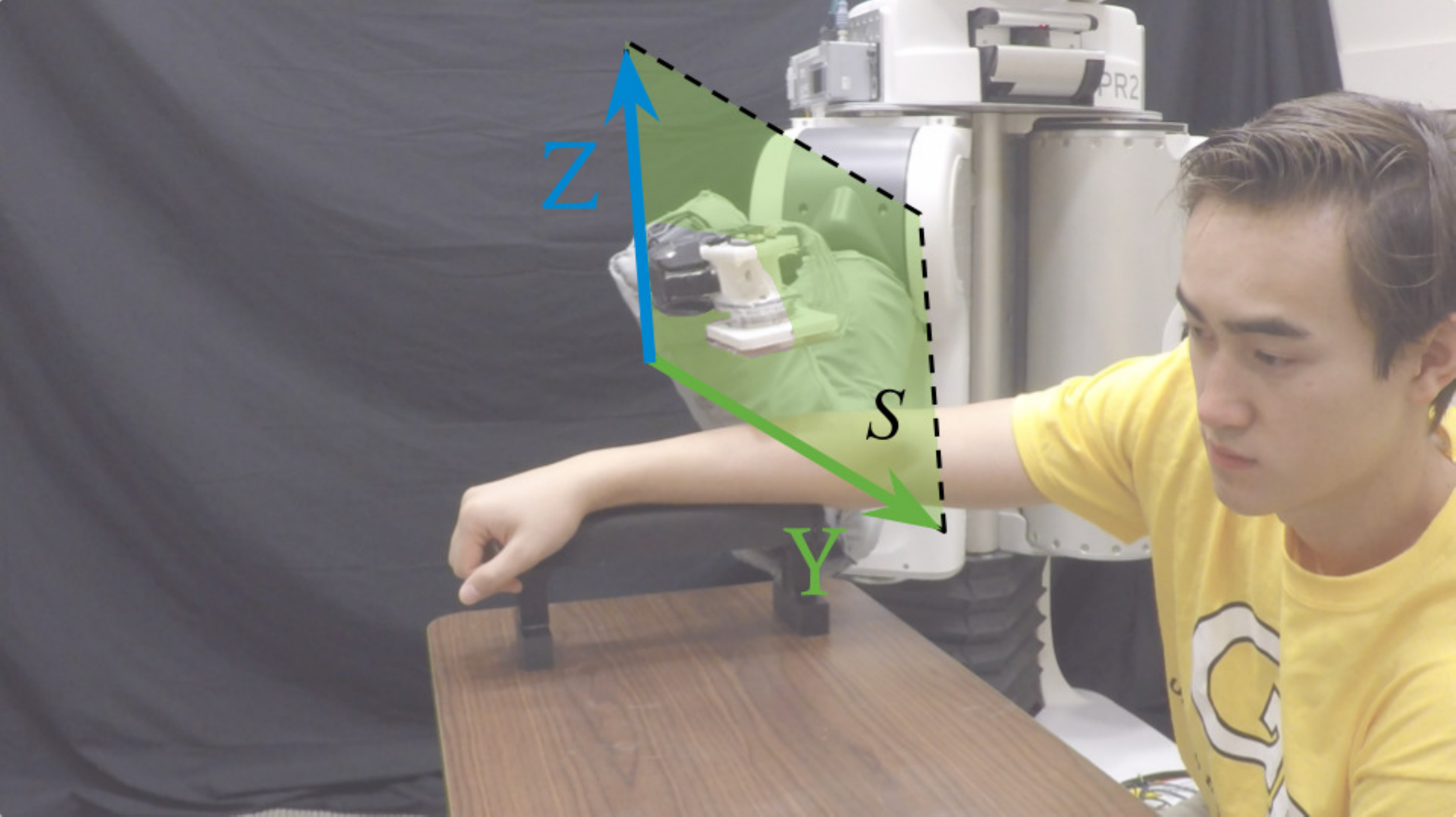}
\includegraphics[width=0.23\textwidth, trim={2cm 3cm 10cm 0cm}, clip]{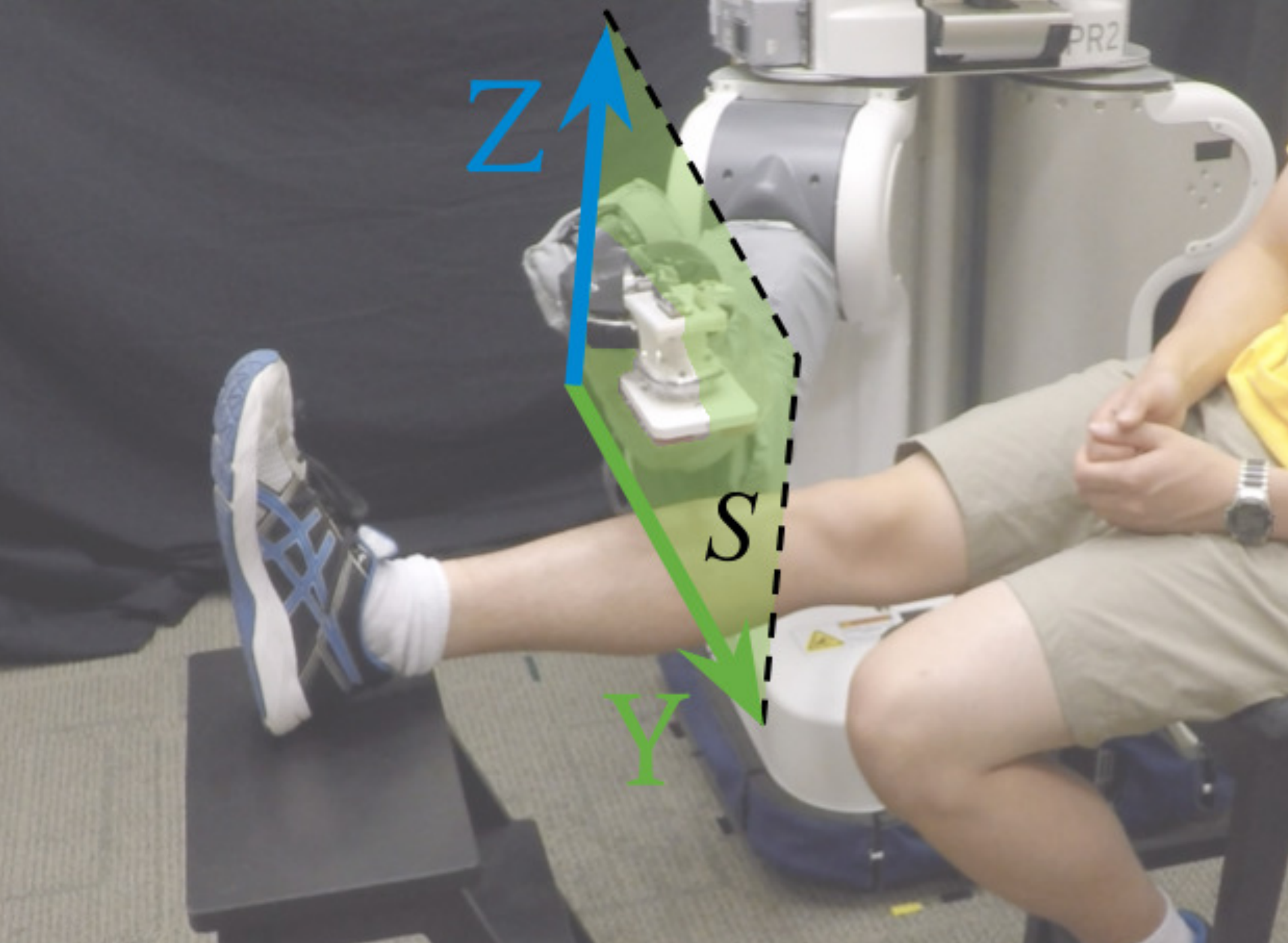}
\caption{\label{fig:datacollection}The participants elevated their arms and legs parallel to the ground during data collection using an armrest and footstool, respectively. The green highlighted region represents the bounded space of target end effector positions used during data collection.}
\vspace{-0.4cm}
\end{figure}

In order to train the network shown in Fig.~\ref{fig:nn}, we collected time-varying capacitance measurements from a single human participant as a PR2 moves the capacitive sensor around both the person's arm and leg. In Section~\ref{sec:evaluation}, we demonstrate how the resulting model can be used by a robot to provide assistance to multiple participants. The PR2 is a general-purpose mobile manipulator, built by Willow Garage, with two 7-DoF back-drivable arms. At each time step $t$ during data collection, we recorded the 50 most recent measurements from the capacitive sensor, $\bm{c}^{}_{t-49:t}$, as well as the relative position of the closest point on the limb and the orientation of the limb's central axis, $\bm{y}^{}_t$.

We collected approximately 630,000 data pairs $(\bm{c}^{}_{t-49:t}, \bm{y}^{}_t)$ distributed across three locations along the arm (wrist, forearm, upper arm), and three locations along the leg (ankle, shin, knee). We sampled data points at a frequency of 100~Hz as the robot moved around a participant's limb, resulting in $\sim$17.5 minutes of data collection per limb location. As depicted in Fig.~\ref{fig:datacollection}, the participant elevated their arm and leg parallel to the ground using an armrest and footstool. Algorithm~\ref{alg:datacollect} details our process for collecting training data with the PR2.

We run Algorithm~\ref{alg:datacollect} at six positions along the arm and leg. Before data collection begins, we provide the robot with a starting position and orientation above the participant's limb, such that $\bm{y}^{}_0 = (0, 0, 0, 0)$. We then randomly select a target position $(p^{}_{T,y}, p^{}_{T,z})$ and orientation $(\theta^{}_{T,y}, \theta^{}_{T,z})$ for the end effector from a uniform distribution such that 
$\bm{y}^{}_T = (p^{}_{T,y}, p^{}_{T,z}, \theta^{}_{T,y}, \theta^{}_{T,z})$ (line 4), where $p^{}_{T,y}\in[-10\text{cm},10\text{cm}]$, $p^{}_{T,z}\in[0\text{cm},15\text{cm}]$, $\theta^{}_{T,y}\in[-\frac{\pi}{8},\frac{\pi}{8}]$, and $\theta^{}_{T,z}\in[-\frac{\pi}{8},\frac{\pi}{8}]$. The bounded region of possible end effector positions used during data collection is depicted in Fig.~\ref{fig:datacollection}. We denote this bounded space of all possible positions and orientations away from the closest point on the limb as $S$.
We then select target velocities (lines 5-6) for the end effector's motion as it moves from $\bm{y}^{}_0$ to $\bm{y}^{}_T$, including translation velocity, $\bm{v}^{}_p$ in cm/s, and rotation velocity, $\bm{v}^{}_\theta$ in radians/s.

At each time step, we record the most recent measurement from the capacitive sensors, $\bm{c}^{}_t$, along with the position of the end effector with respect to the closest point on the limb and orientation of the end effector with respect to the limb's central axis, $\bm{y}^{}_t$ (lines 9-12). 
The robot then takes an action along a linear trajectory towards the target end effector position and orientation, with velocities $\bm{v}^{}_p$ and $\bm{v}^{}_\theta$, respectively. Finally, we repeat this entire process for a total of $N=500$ iterations (line 3), resulting in a dataset $\mathcal{D}$ of pairings $(\bm{c}^{}_{t-h+1:t}, \bm{y}^{}_t)$ for training our model.

\subsection{Control}
\label{sec:control}

\newlength{\textfloatsepsave}
\setlength{\textfloatsepsave}{\textfloatsep}
\setlength{\textfloatsep}{10pt}

\begin{algorithm}[t]
\caption{Human Limb Contour Following}\label{alg:control}
\begin{algorithmic}[1]
\State \textbf{input:} estimation model $f$, control policy $u$, desired position and orientation for end effector $\bm{y}^{}_{desired}$, time steps between action updates $\tau$.
\State $t \gets 0$.
\While{force $<$ 10 N and end of limb not reached}
\State Observe capacitance measurements $\bm{c}^{}_t\in\mathbb{R}^6$.
\If{$t \bmod \tau = 0$}
\State Compute error $e(t)$ using model $f(\bm{c}^{}_{t-h+1:t})$.
\State Compute action $u(t)$ given control policy in (\ref{eq:controller}).
\State Execute $u(t)$.
\EndIf
\State $t \gets t+1$.
\EndWhile
\end{algorithmic}
\end{algorithm}

Throughout our experiments in Section~\ref{sec:evaluation}, we control the PR2 to follow the path and contours of a person's limb using a high level Cartesian controller. Our controller uses the TRAC-IK\footnote{TRAC-IK: \url{https://traclabs.com/projects/trac-ik/}} inverse kinematics library to provide joint-level input to the low-level PID controllers at each actuator in the PR2's arm. We used low PID gains at each joint so that the robot's arms were compliant for participant safety. In addition, we implemented a force threshold monitor, which would halt robot movement if the end effector made contact with a participant during an experiment and forces measured at the force/torque sensor exceeded 10~N.

\begin{figure}
\centering
\includegraphics[width=0.48\textwidth, trim={2cm 3cm 3cm 3cm}, clip]{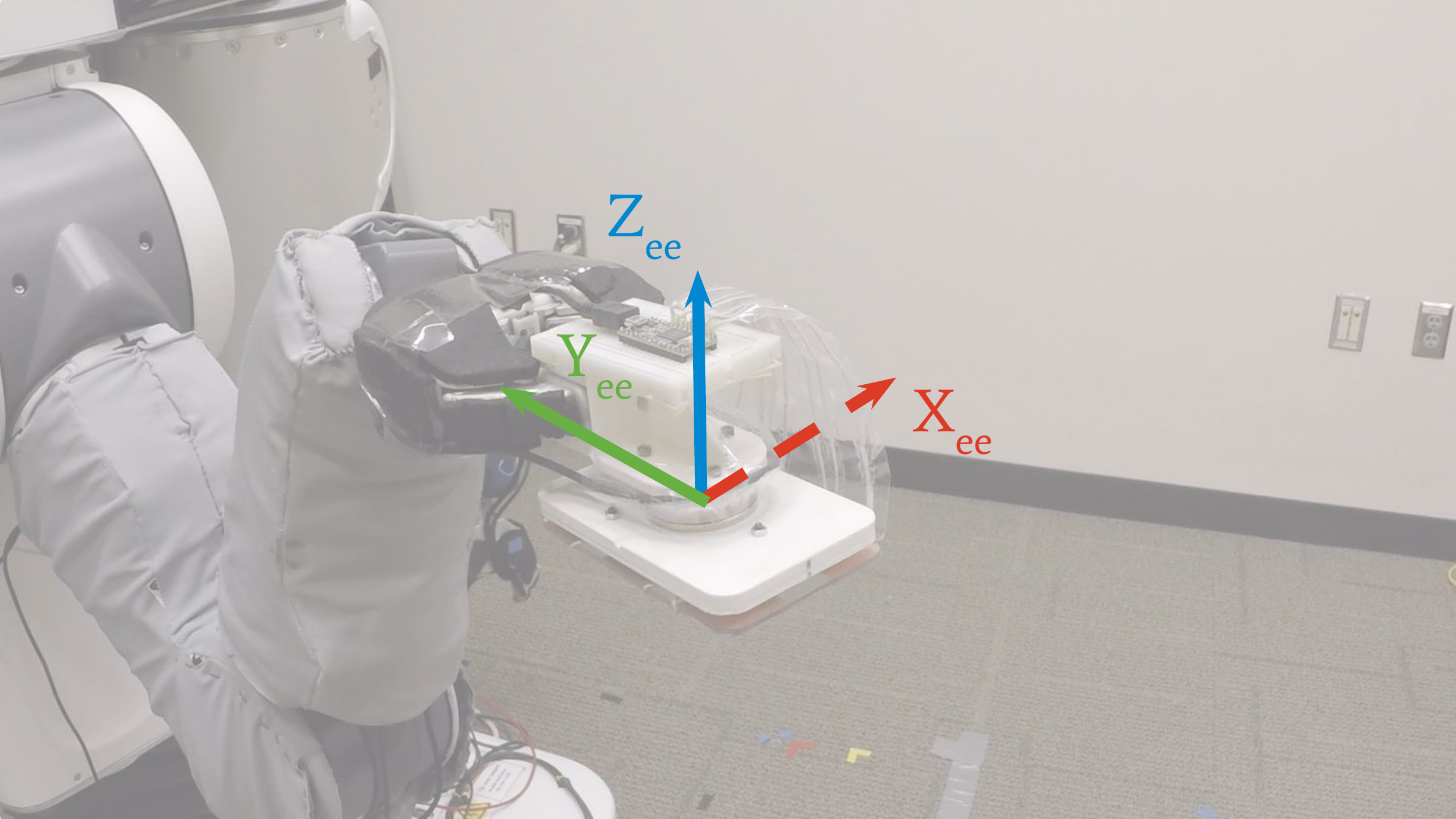}
\vspace{-0.4cm}
\caption{\label{fig:endeffector}The local coordinate frame of the end effector. During our evaluations with human participants, the end effector traverses along the $X_{ee}$ axis at a rate of 2~cm/s.}
\vspace{-0.4cm}
\end{figure}

During execution, the controller commanded the PR2's end effector to move 2~cm/s along the X-axis of the end effector, $X_{ee}$, (see Fig.~\ref{fig:endeffector}), approximately along the central axis of the limb. 
At a time step $t$, the robot used our trained model to estimate the position of the closest point on a person's limb and the orientation of the limb's central axis, $\bm{\hat{y}}^{}_t = f(\bm{c}^{}_{t-h+1:t})$. The controller then computed actions for the robot, $u(t) = (u^{}_y(t), u^{}_z(t), u^{}_{\theta_y}(t), u^{}_{\theta_z}(t))$, representing discrete translational movements $(u^{}_y(t), u^{}_z(t))$ and rotational motions $(u^{}_{\theta_y}(t), u^{}_{\theta_z}(t))$ towards an offset from the closest point on the human limb, $\bm{y}^{}_{desired} = (p^{}_y, p^{}_z, \theta^{}_y, \theta^{}_z)$. Since our model estimates are with respect to a coordinate frame at the closest point on the person's limb, the robot transforms these actions into the end effector's frame of reference prior to executing the actions.
We can also change $\bm{y}^{}_{desired}$ to adapt to the robot's task at hand. For example, when our robot is providing dressing assistance (Section~\ref{sec:dressing}), we define $\bm{y}^{}_{desired} = (0, 5, 0, 0)$, such that the end effector stays 5~cm above a participant's limb, with the same orientation as the limb. However, when the robot is providing bathing assistance (Section~\ref{sec:bathing}), we set $\bm{y}^{}_{desired} = (0, 1, 0, 0)$ so that the wet washcloth held by the robot makes direct contact with a person's skin. We define our PD controller as,

\begin{equation}
    \begin{aligned}
        u(t)  =  \bm{K}^{}_p e(t) + \bm{K}^{}_d \dot{e}(t)\\
    \end{aligned} \label{eq:controller}
\end{equation}
where
\begin{equation}
    \begin{aligned}
        e(t)  &=   \bm{y}^{}_{desired} - \bm{\hat{y}}^{}_t\\
        &=   \bm{y}^{}_{desired} - f(\bm{c}^{}_{t-h+1:t}).
    \end{aligned} \label{eq:error}
\end{equation}

\setlength{\textfloatsep}{\textfloatsepsave}

$e(t)$ is the tracking error, whereas $\bm{K}^{}_p$ and $\bm{K}^{}_d$ are diagonal matrices for the proportional and derivative gains, respectively. We tuned the controller to produce smooth motion while following the contours of a person's limb, resulting in $\bm{K}^{}_p = \text{diag}(0.025, 0.025, 0.1, 0.1)$ and $\bm{K}^{}_d = \text{diag}(0.0125, 0.0125, 0.025, 0.025)$. When evaluating the robot's responsiveness to human motion in Section~\ref{sec:dressing}, we changed the proportional gains of our controller to $\bm{K}^{}_p = \text{diag}(0.2, 0.2, 0.1, 0.1)$.

Algorithm~\ref{alg:control} details the control process we used throughout our evaluations in Section~\ref{sec:evaluation}. During an evaluation trial, the robot continually samples capacitance measurements and updates its actions according to the control policy, $u$, until the trial completes. The robot observes measurements from the capacitive sensor at 100~Hz (line 4 of Algorithm~\ref{alg:control}) and runs the high-level Cartesian controller at a rate of 10~Hz ($\tau=10$). The high-level Cartesian controller uses an updated action to produce new target actuator angles for the low-level PID controllers (line 8). We marked the end of an evaluation trial (line 3) whenever the applied forces measured at the end effector exceeded 10 N, or when the robot's end effector had traversed a participant's entire limb, such as starting above a participant's hand and navigating up to the shoulder.

\section{Evaluation}
\label{sec:evaluation}

In the following sections, we describe how our capacitive sensing approach can enable a robot to assist people with two everyday tasks: dressing, during which a robot pulls the sleeve of a hospital gown onto a participant's arm, and bathing, during which the robot uses a wet cloth to wipe down the surface of a person's body.

To evaluate our capacitive sensor and trained model, we conducted a study with four able-bodied human participants (two females and two males) with informed consent from all participants and approval from the Georgia Institute of Technology Institutional Review Board (IRB). We recruited participants to meet the following inclusion/exclusion criteria: $\geq$ 18 years of age; fluent in written and spoken English; and have not been diagnosed with ALS or other forms of motor impairments. Participant ages ranged from 18 to 27 years with arm lengths between 57-59~cm and leg lengths between 83-90~cm.
For each of the assistive dressing and bathing scenarios detailed below, we conducted two trials between the robot and each participant. Our study and experiments can be found in the supplementary video.

\subsection{Traversing the Human Arm}

\begin{figure*}
\centering
\includegraphics[width=0.24\textwidth, trim={10cm 5cm 5cm 0cm}, clip]{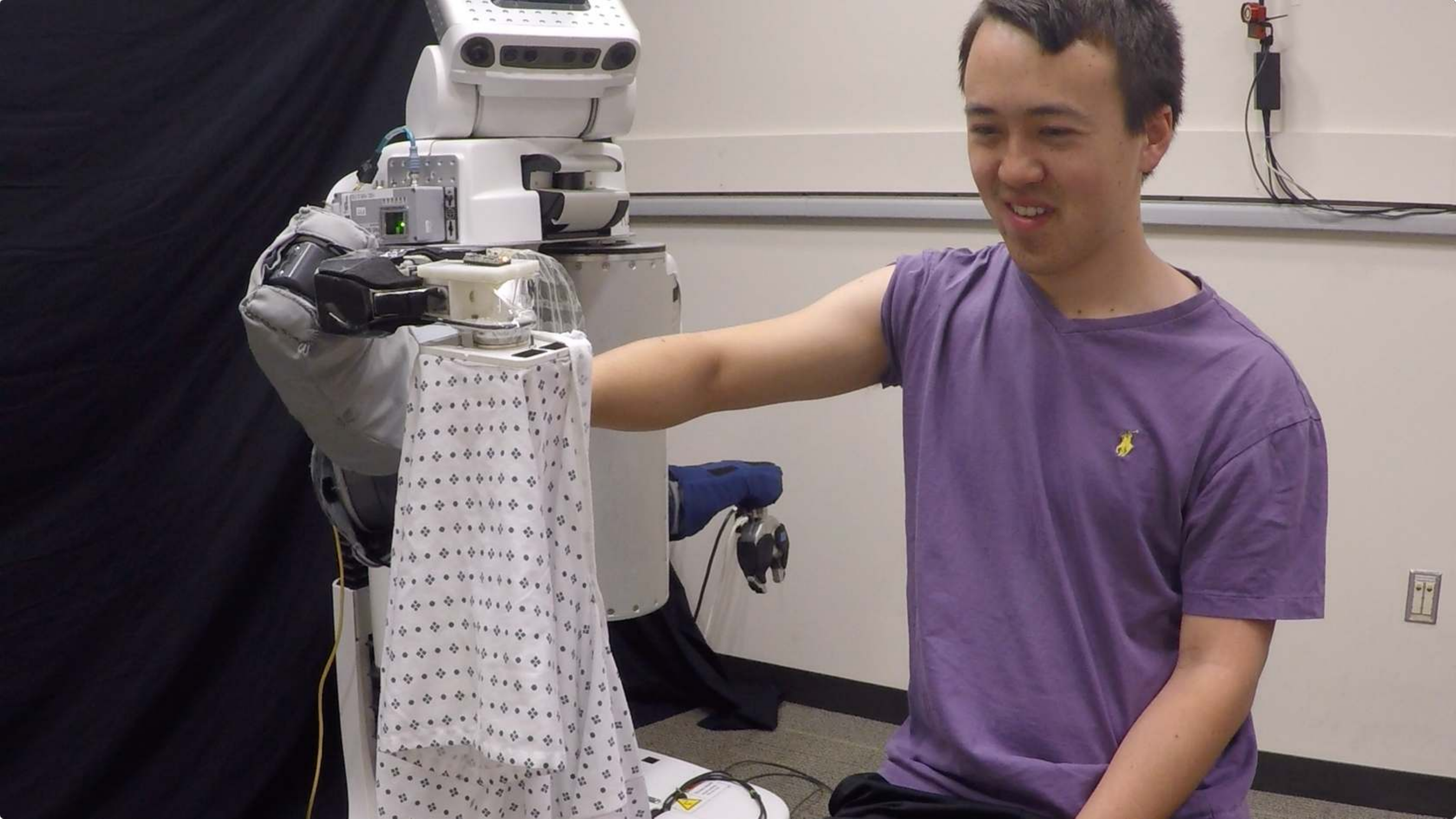}
\includegraphics[width=0.24\textwidth, trim={10cm 5cm 5cm 0cm}, clip]{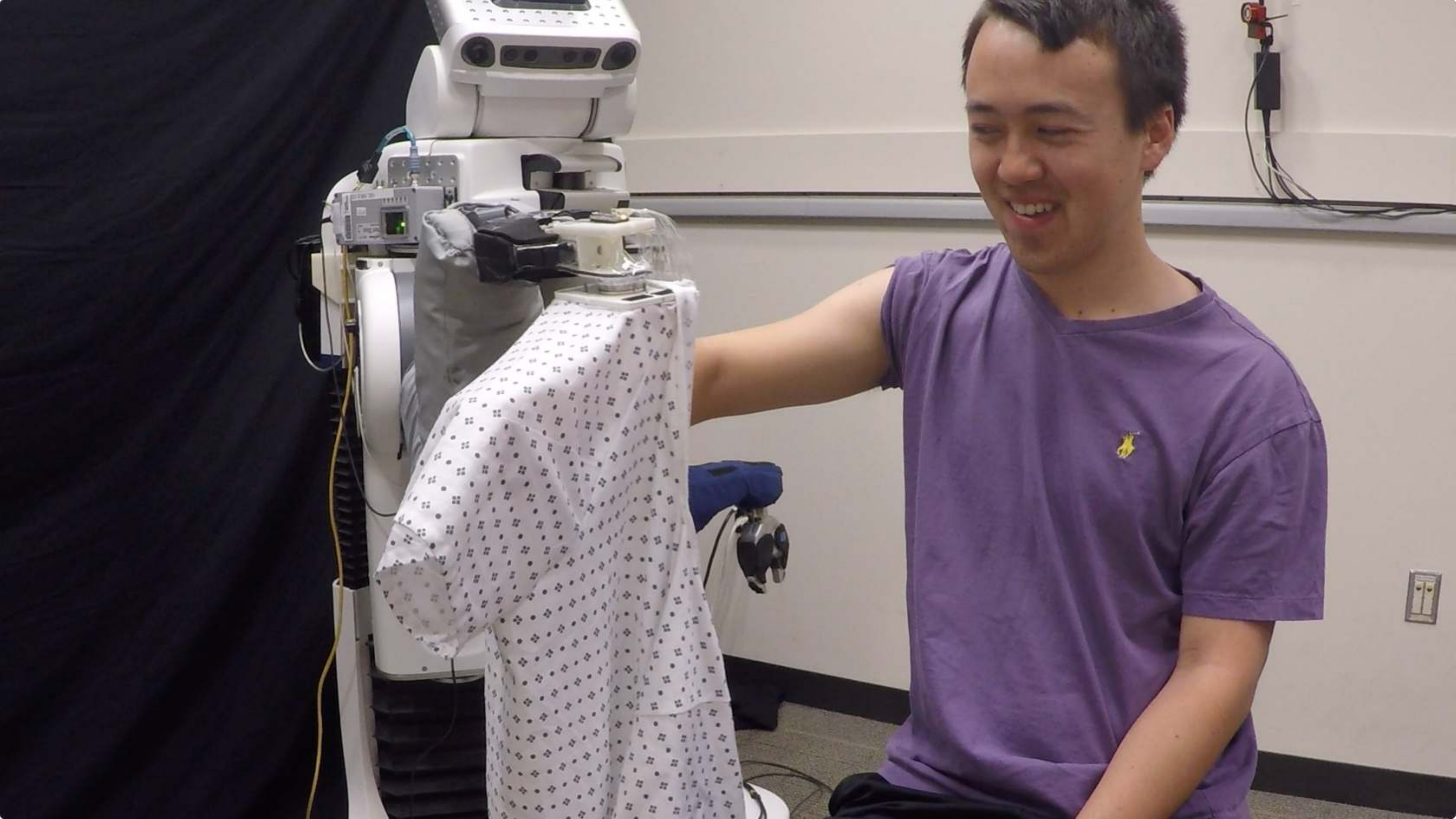}
\includegraphics[width=0.24\textwidth, trim={10cm 5cm 5cm 0cm}, clip]{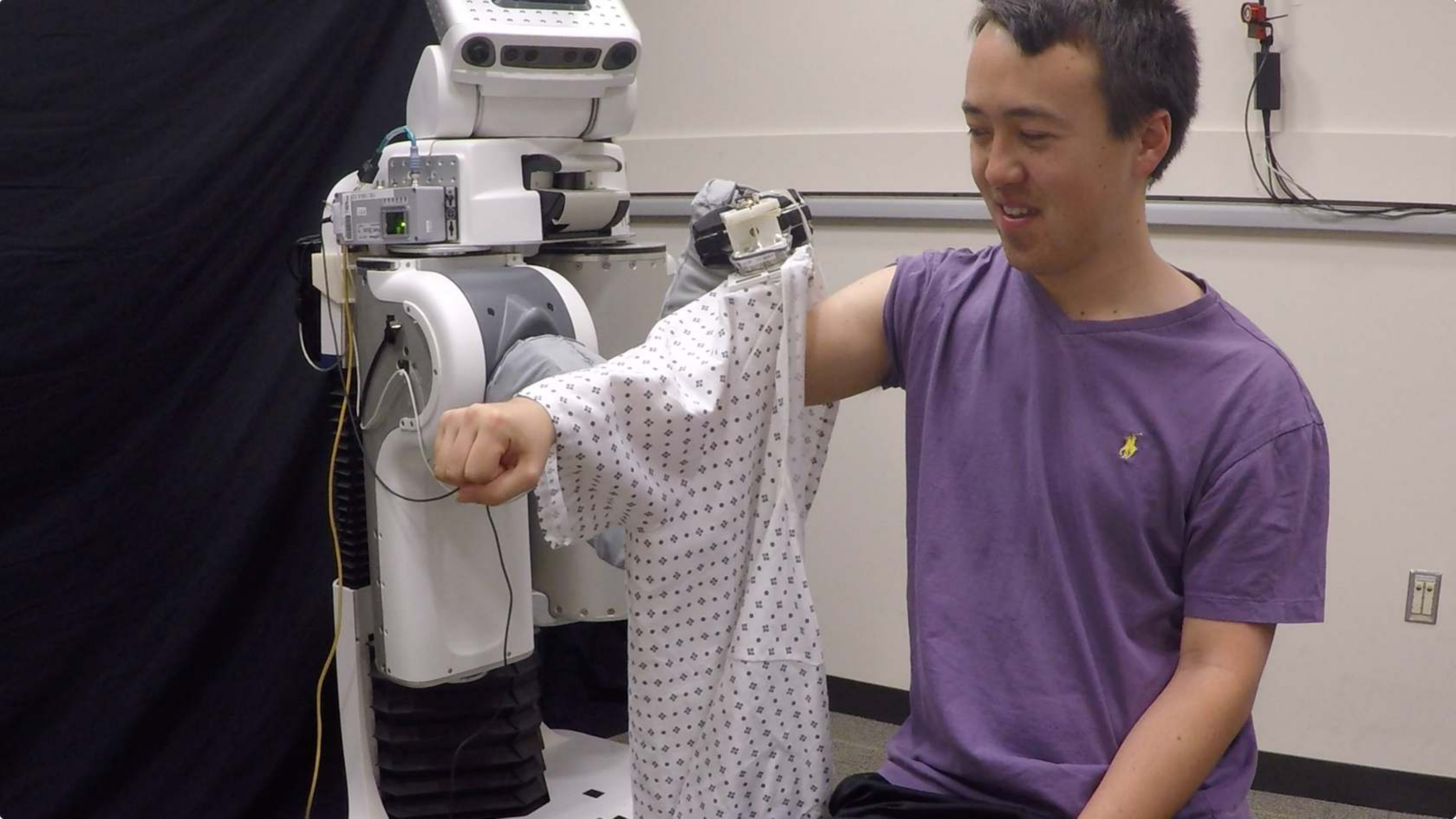}
\includegraphics[width=0.24\textwidth, trim={10cm 5cm 5cm 0cm}, clip]{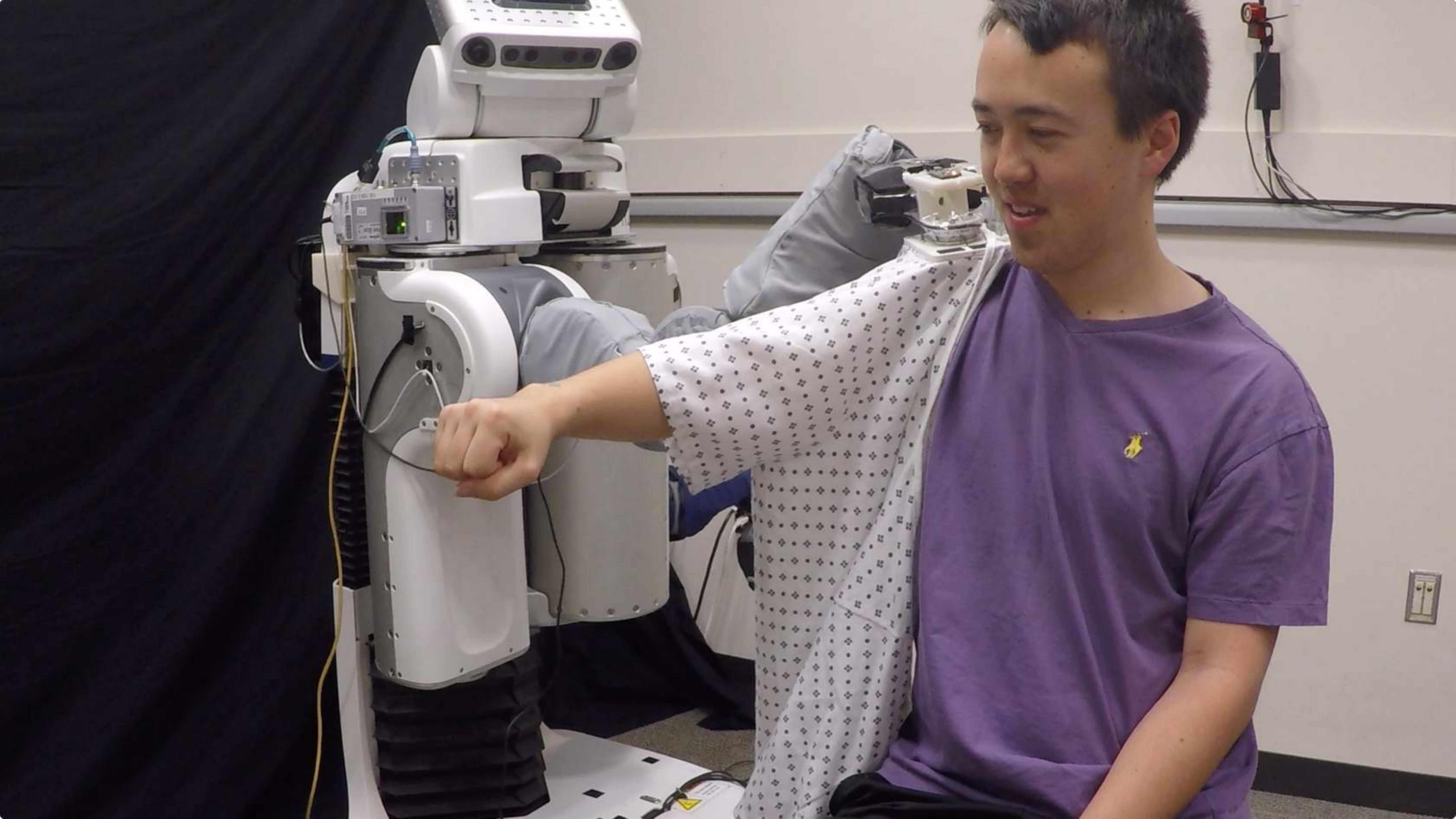}
\vspace{-0.2cm}
\caption{\label{fig:elbow}The PR2 using capacitive sensing to follow the path of a participant's arm and pull on the sleeve of a hospital gown. Participant's held their arm parallel to the ground plane with a 30 degree elbow bend.}
\vspace{-0.2cm}
\end{figure*}

\begin{figure*}
\centering
\includegraphics[width=0.24\textwidth, trim={10cm 0cm 5cm 0cm}, clip]{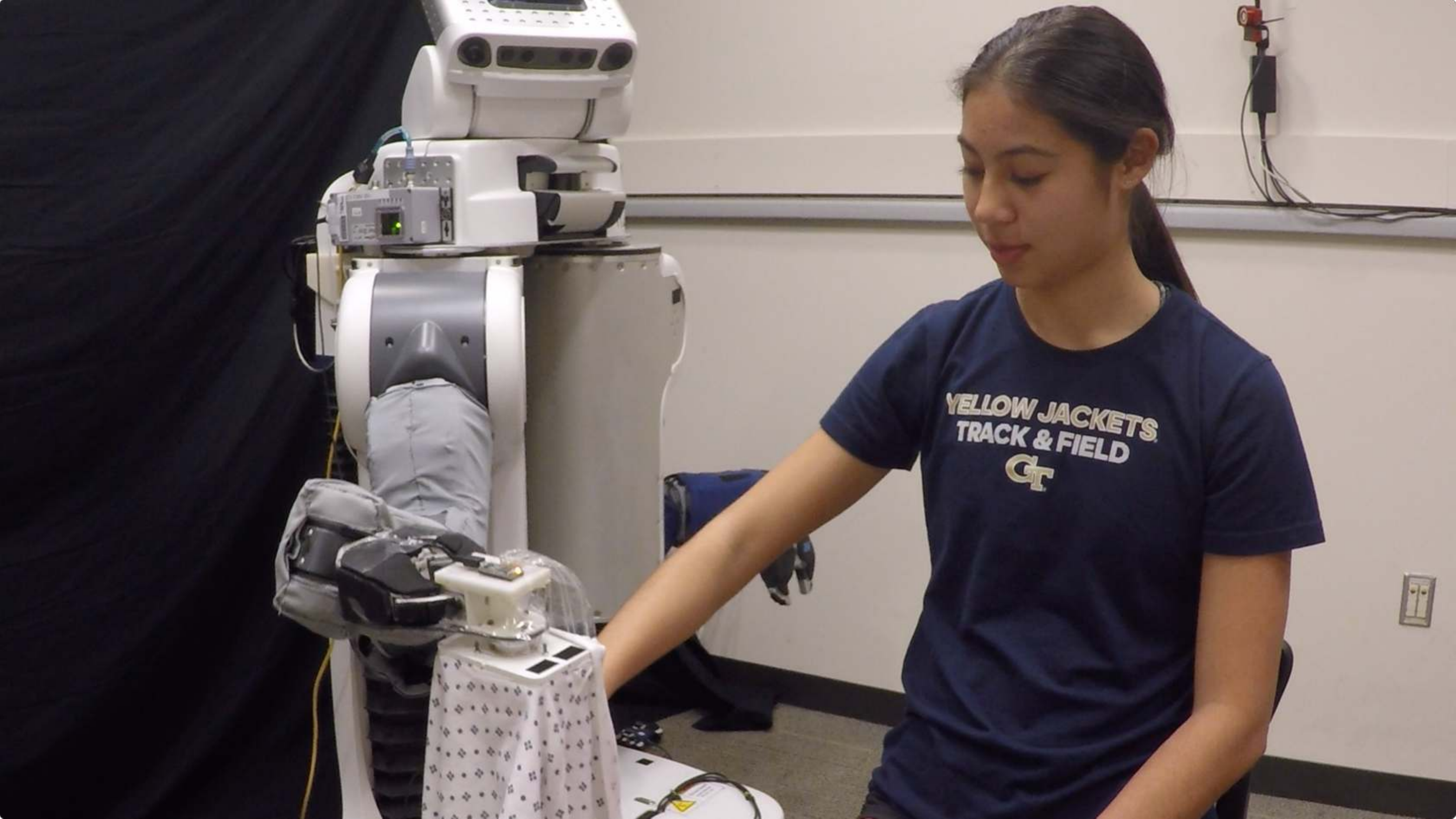}
\includegraphics[width=0.24\textwidth, trim={10cm 0cm 5cm 0cm}, clip]{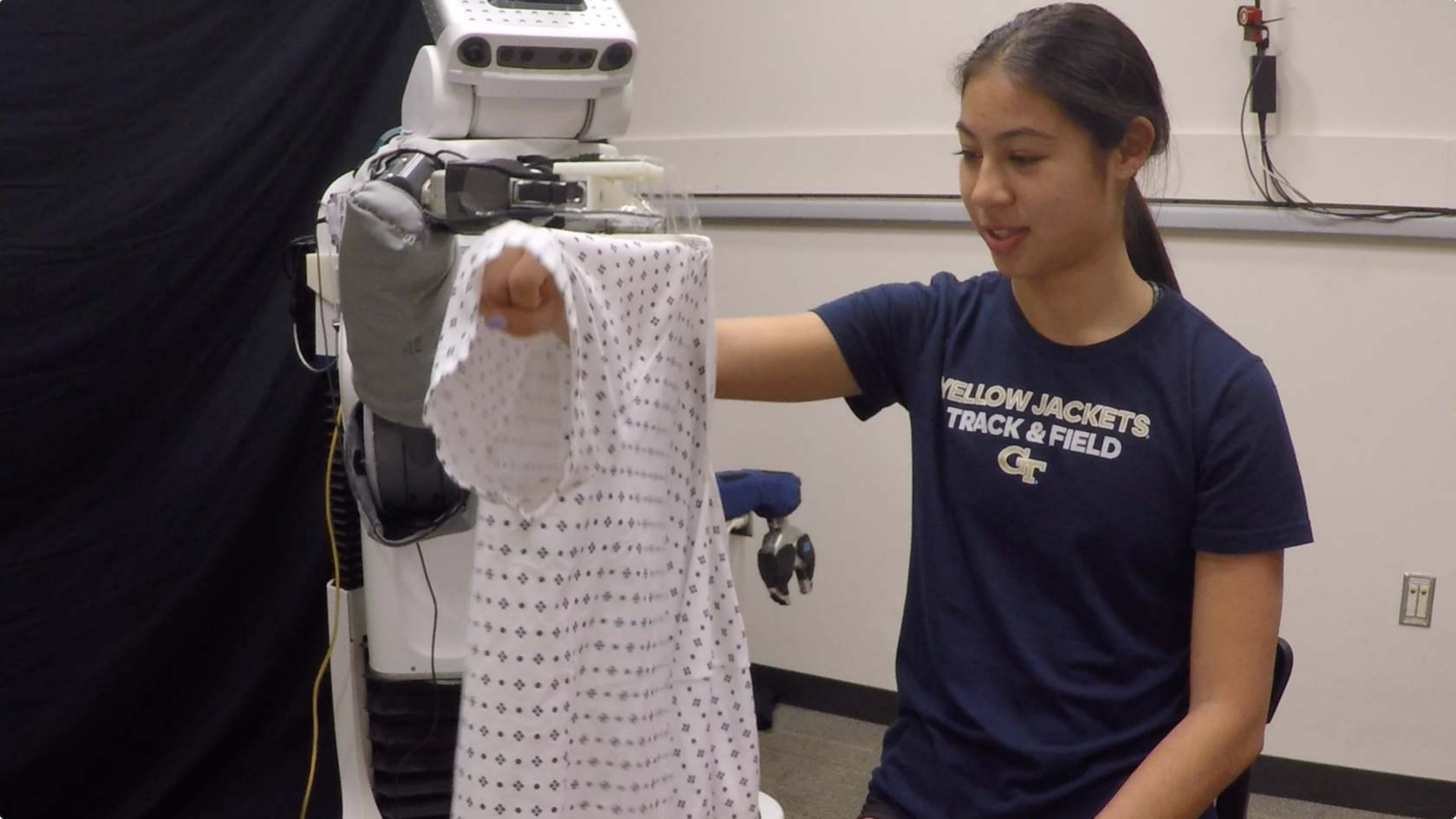}
\includegraphics[width=0.24\textwidth, trim={10cm 0cm 5cm 0cm}, clip]{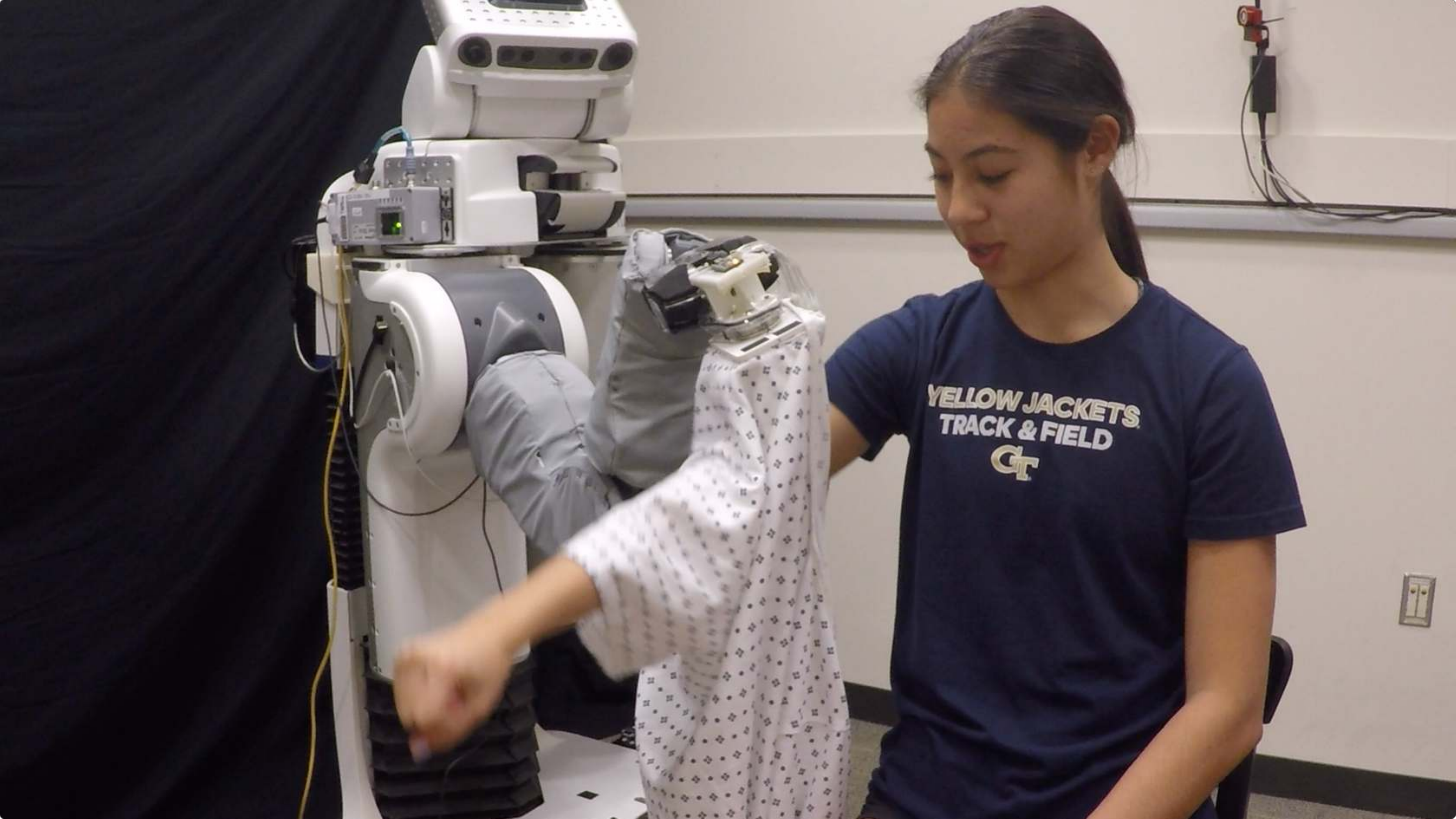}
\includegraphics[width=0.24\textwidth, trim={10cm 0cm 5cm 0cm}, clip]{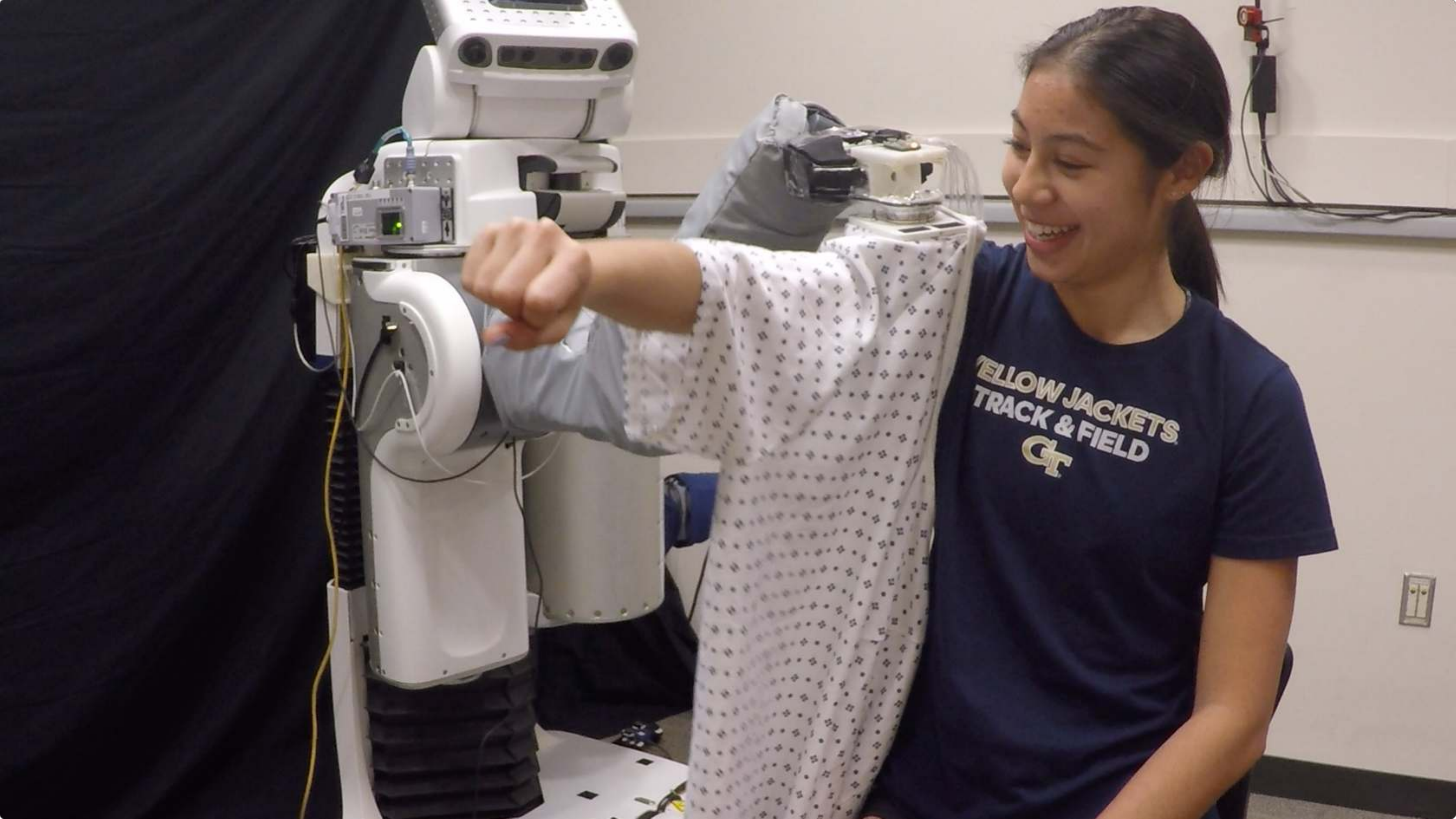}
\vspace{-0.2cm}
\caption{\label{fig:vertical_movement}The PR2 using capacitive sensing to track vertical arm motion. The PR2's end effector raised and lowered in response to the participant's movement, while also performing pitch rotations to match the orientation of the arm (see third figure in sequence).}
\vspace{-0.4cm}
\end{figure*}

\begin{figure}
\centering
\includegraphics[width=0.48\textwidth, trim={1cm 19.5cm 2.5cm 2.7cm}, clip]{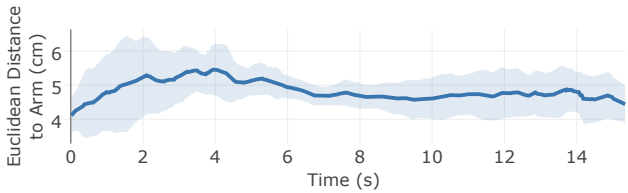}
\vspace{-0.6cm}
\caption{\label{fig:distance}Euclidean distance between the robot's end effector and a participant's arm. Given the reflective markers on a participant's joints, we define 3D line segments representing the forearm and upper arm, and we compute distance as the shortest path between the end effector and these line segments. Distance results are averaged across trials from all participants. Shaded regions represent one standard deviation.}
\vspace{-0.4cm}
\end{figure}

We began by evaluating the robot's ability to use capacitive sensing to follow the contours of a participant's arm and dress the sleeve of a hospital gown (Fig.~\ref{fig:elbow}). 


During a dressing trial, participants held their arm up parallel to the ground with their elbow bent at a 30 degree rotation. The PR2 held a hospital gown in its right end effector and began with the capacitive sensor starting $\sim$5~cm above the center of a participant's hand. The robot then used capacitance measurements to follow the contours of a participant's arm, according to Algorithm~\ref{alg:control}. We conducted two dressing trials per participant, totalling eight trials across all participants. Fig.~\ref{fig:elbow} shows the PR2 using the capacitive sensor to follow the path of a participant's arm and pull the sleeve of a hospital gown fully up to the shoulder. Notice that once the robot's end effector reached a participant's elbow, the capacitive sensor sensed a change in orientation between the forearm and upper arm. This resulted in the end effector rotating 30 degrees around the bent elbow to match the orientation of the upper arm.
For dressing, we classified a trial as successful if the robot's end effector reached the top of a participant's shoulder with the sleeve around the upper arm. Overall, the robot succeeded in pulling the gown up to the top of a participant's shoulder in all 8 trials.

In order to obtain the ground truth position and orientation of each participant's arm for post-hoc comparison, we placed small infrared reflective markers on a participant's wrist, elbow, and shoulder. We used the PR2's head-mounted Microsoft Kinect depth sensor to record these 3D joint positions, yet these positions were not provided to the robot during a trial. After each trial was completed, we computed the distance between the robot's end effector and the top of a participant's arm at each time step using 3D line segments between these joint positions. Fig.~\ref{fig:distance} depicts the Euclidean distance and standard deviation between the robot's end effector and a participant's arm, averaged across all participants. We observed that, on average, the robot's end effector remained within 4 to 6~cm from the surfaces of the participants' arms during the dressing trials.

\subsection{Adapting to Arm Motion}
\label{sec:dressing}

\begin{figure}
\centering
\includegraphics[width=0.48\textwidth, trim={1cm 19.5cm 2.5cm 2.7cm}, clip]{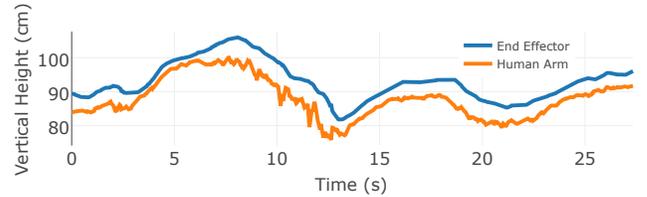}
\vspace{-0.6cm}
\caption{\label{fig:vertical_distance}Vertical position of the robot's end effector and human arm as a participant moves his/her arm vertically during dressing. For clarity, this plot shows a representative trial from a single participant performing vertical arm movements, as seen in Fig.~\ref{fig:vertical_movement}. The position of the end effector is measured from forward kinematics while the vertical position of a participant's arm is estimated using the capacitive sensor and our neural network model.}
\vspace{-0.2cm}
\end{figure}

We also investigated the robot's ability to use multidimensional capacitive sensing to adapt to human motion during robot-assisted dressing. We first instructed participants to perform vertical arm movements by tilting their arm during dressing, such that their hand remained within 20~cm above or below the starting pose. Participants again started with their arm horizontal to the ground plane, with a 30 degree bend in their elbow.
Fig.~\ref{fig:vertical_movement} shows a sequence of images depicting how the robot adjusted to the vertical motions of a participant. In addition, Fig.~\ref{fig:vertical_distance} displays how the vertical position of the end effector rose and lowered to match the estimated height of a participant's arm during a single representative dressing trial. 
Despite arm motions, the robot used capacitive sensing with feedback control to remain on average 4.9~cm above the participant's arm throughout the entire dressing process.
We observed that these results were consistent for all participants as the robot successfully pulled the gown up to each participant's shoulder across all 8 trials.

\begin{figure}
\centering
\includegraphics[width=0.48\textwidth, trim={1cm 19.5cm 2.5cm 2.7cm}, clip]{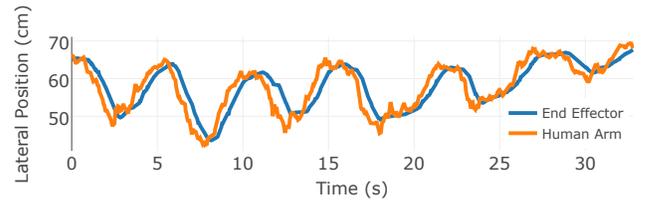}
\vspace{-0.6cm}
\caption{\label{fig:lateral_distance}Lateral position (along the Y-axis) of the end effector and human arm as a participant shifts his/her arm left to right during dressing. This plot displays a representative trial from a single participant. The lateral position of the end effector is measured from forward kinematics whereas the position of a participant's arm is estimated using the capacitive sensor and our neural network model.}
\vspace{-0.4cm}
\end{figure}

\begin{figure*}
\centering
\includegraphics[width=0.24\textwidth, trim={20cm 0cm 0cm 5cm}, clip]{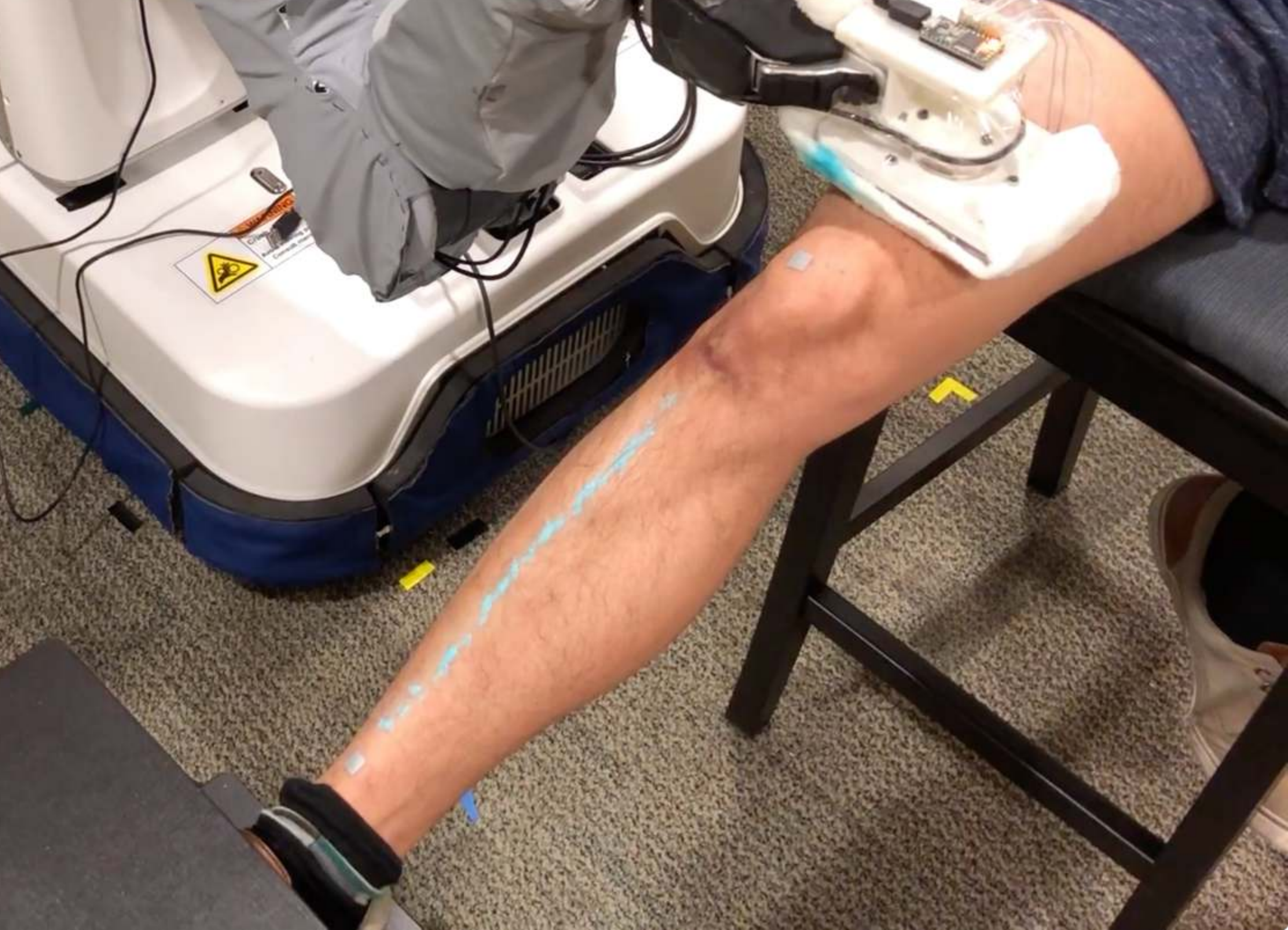}
\includegraphics[width=0.24\textwidth, trim={20cm 0cm 0cm 5cm}, clip]{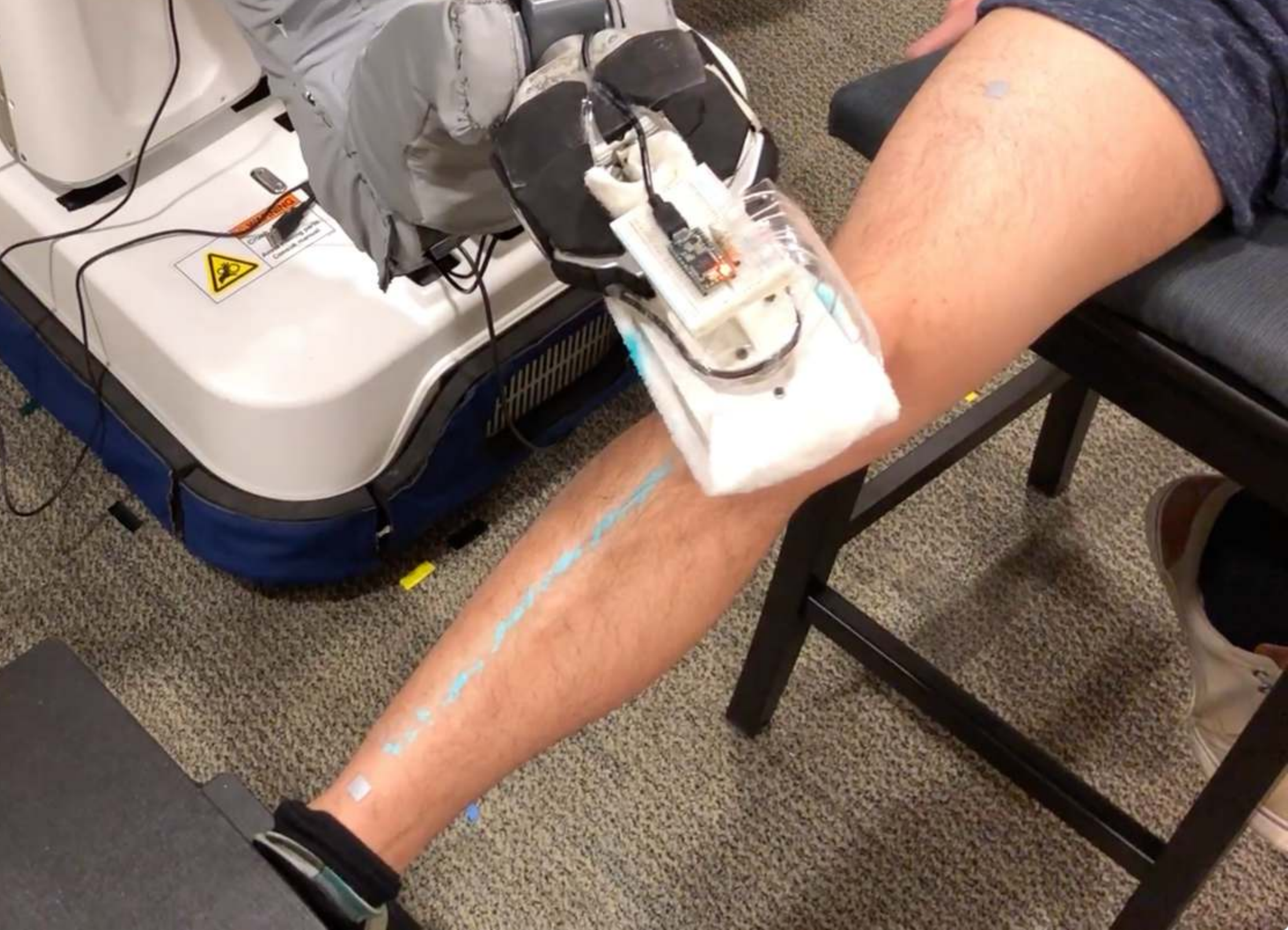}
\includegraphics[width=0.24\textwidth, trim={20cm 0cm 0cm 5cm}, clip]{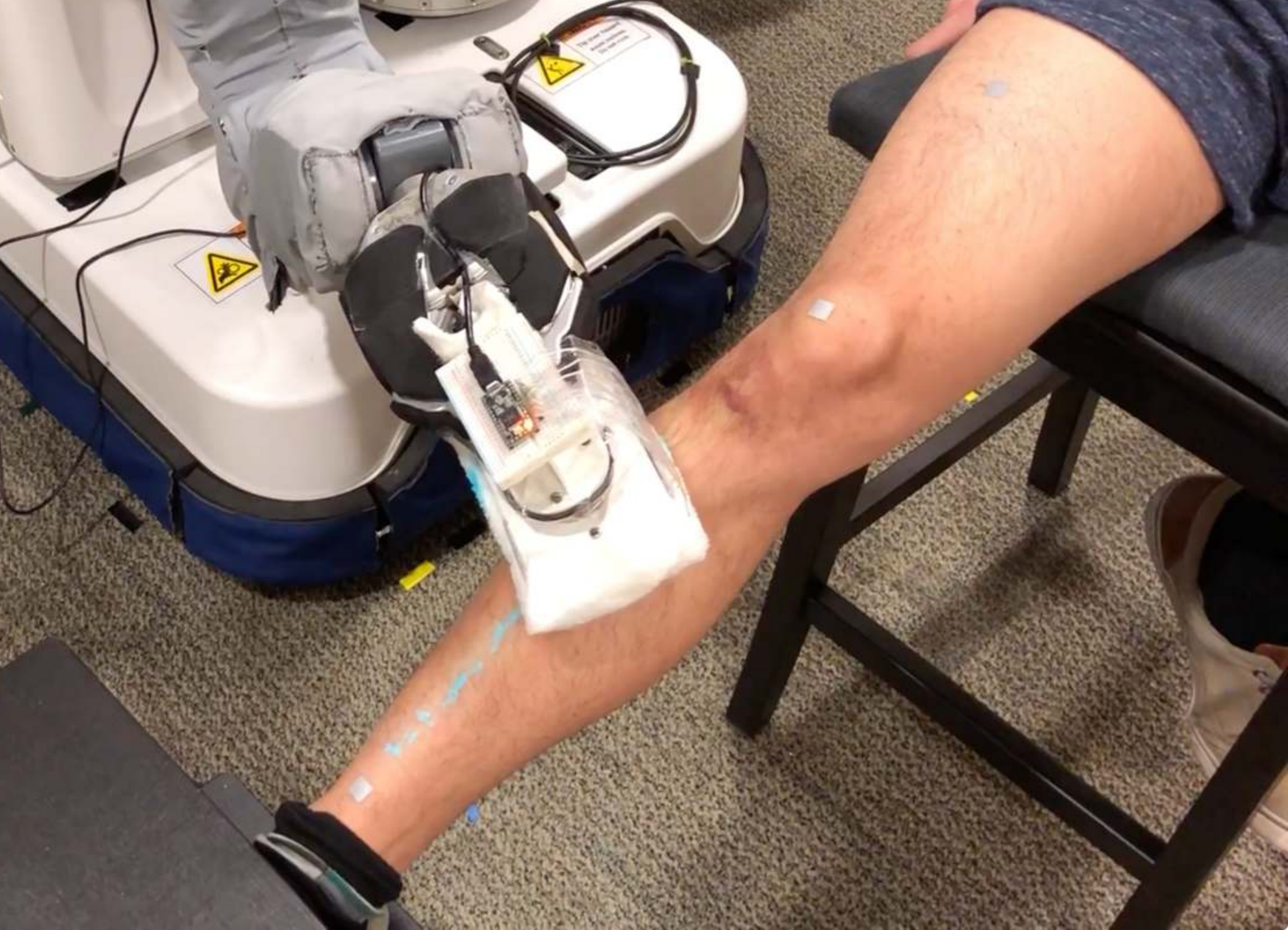}
\includegraphics[width=0.24\textwidth, trim={20cm 0cm 0cm 5cm}, clip]{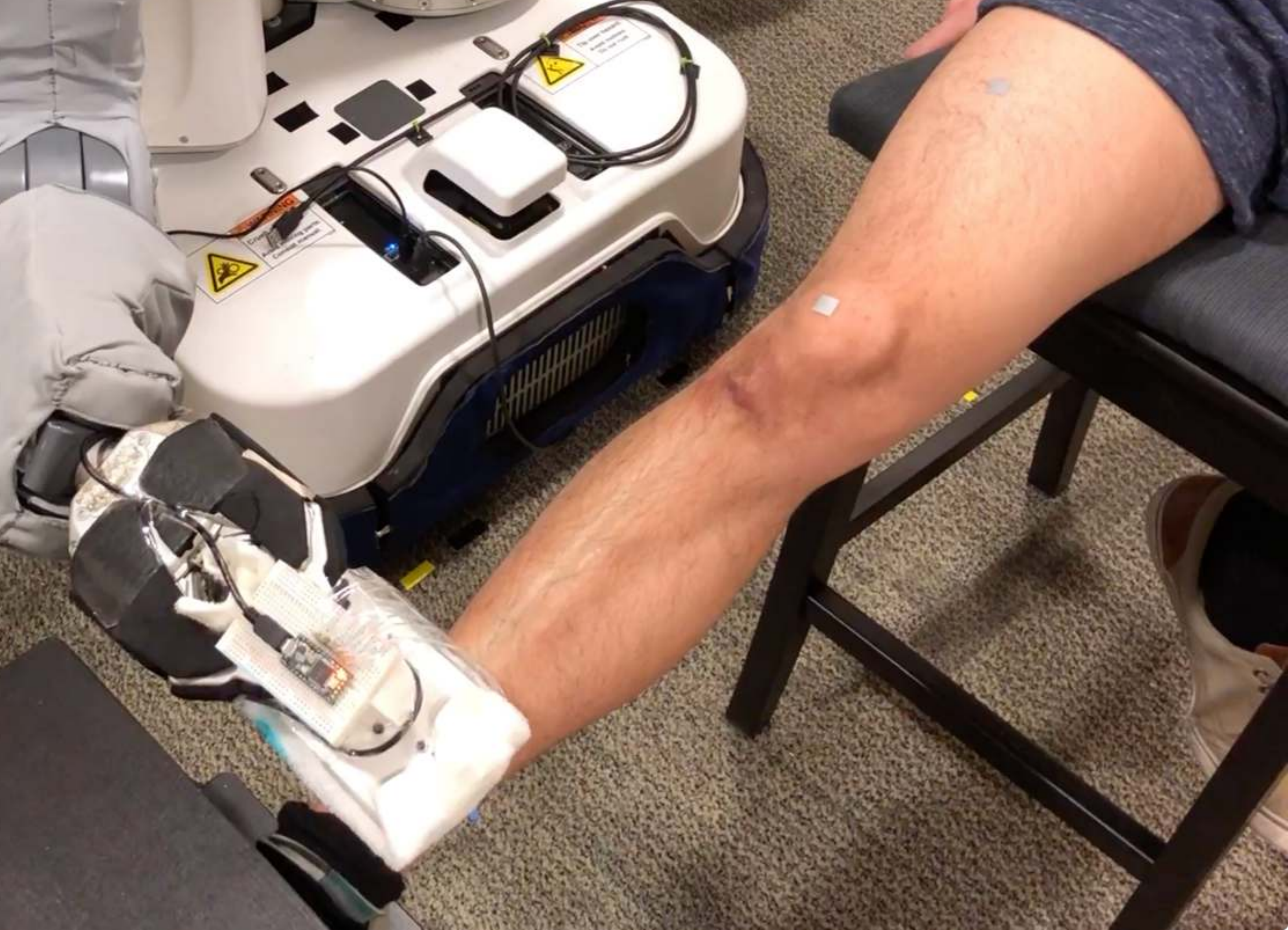}
\vspace{-0.2cm}
\caption{\label{fig:leg_wiping}The PR2 uses capacitive sensing to follow the path of a participant's leg and navigate around a 30 degree knee bend while wiping blue powder off of a participant's lower leg with a wet washcloth.}
\vspace{-0.2cm}
\end{figure*}

\begin{figure*}
\centering
\includegraphics[width=0.24\textwidth, trim={10cm 0cm 5cm 0cm}, clip]{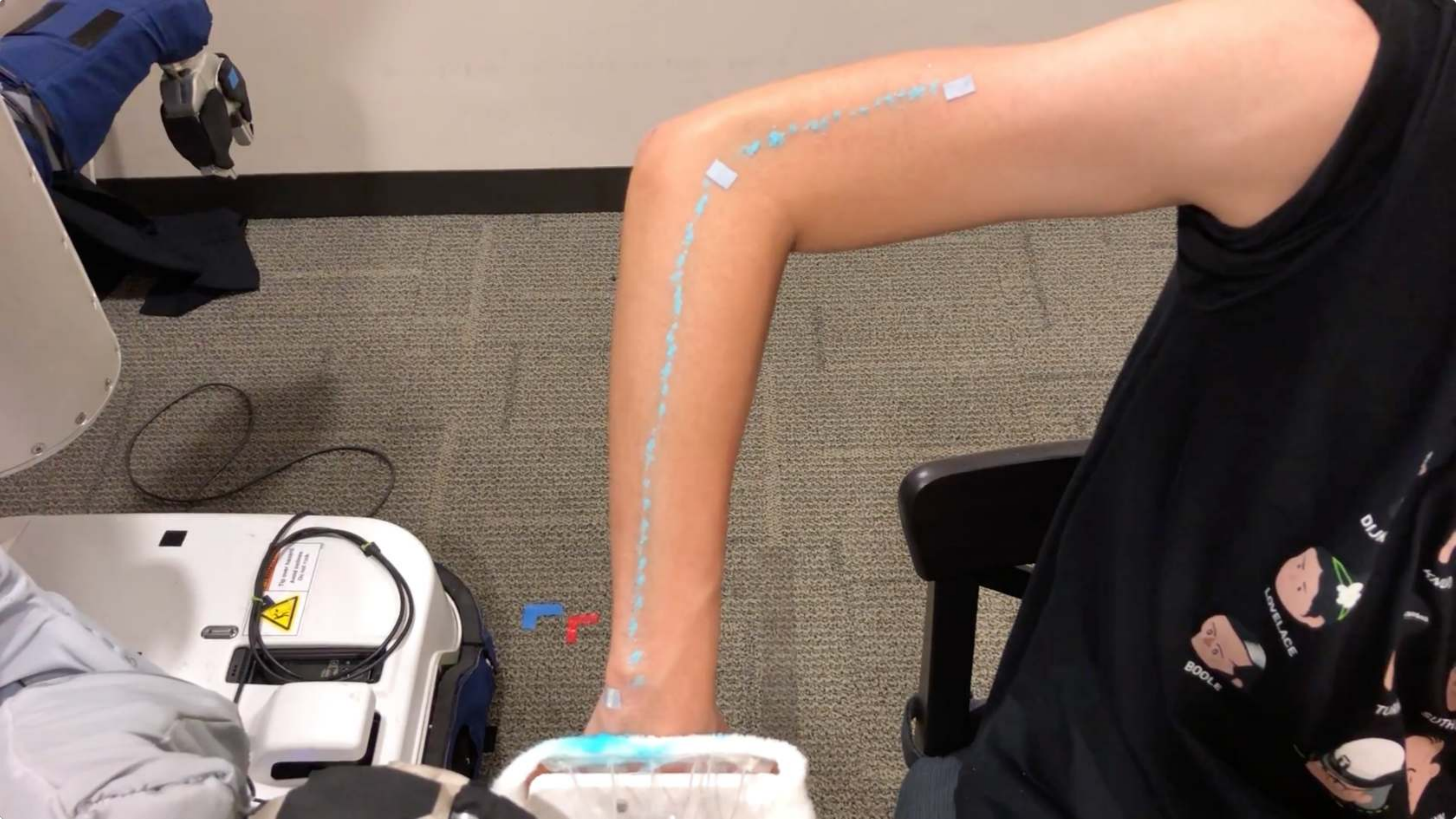}
\includegraphics[width=0.24\textwidth, trim={10cm 0cm 5cm 0cm}, clip]{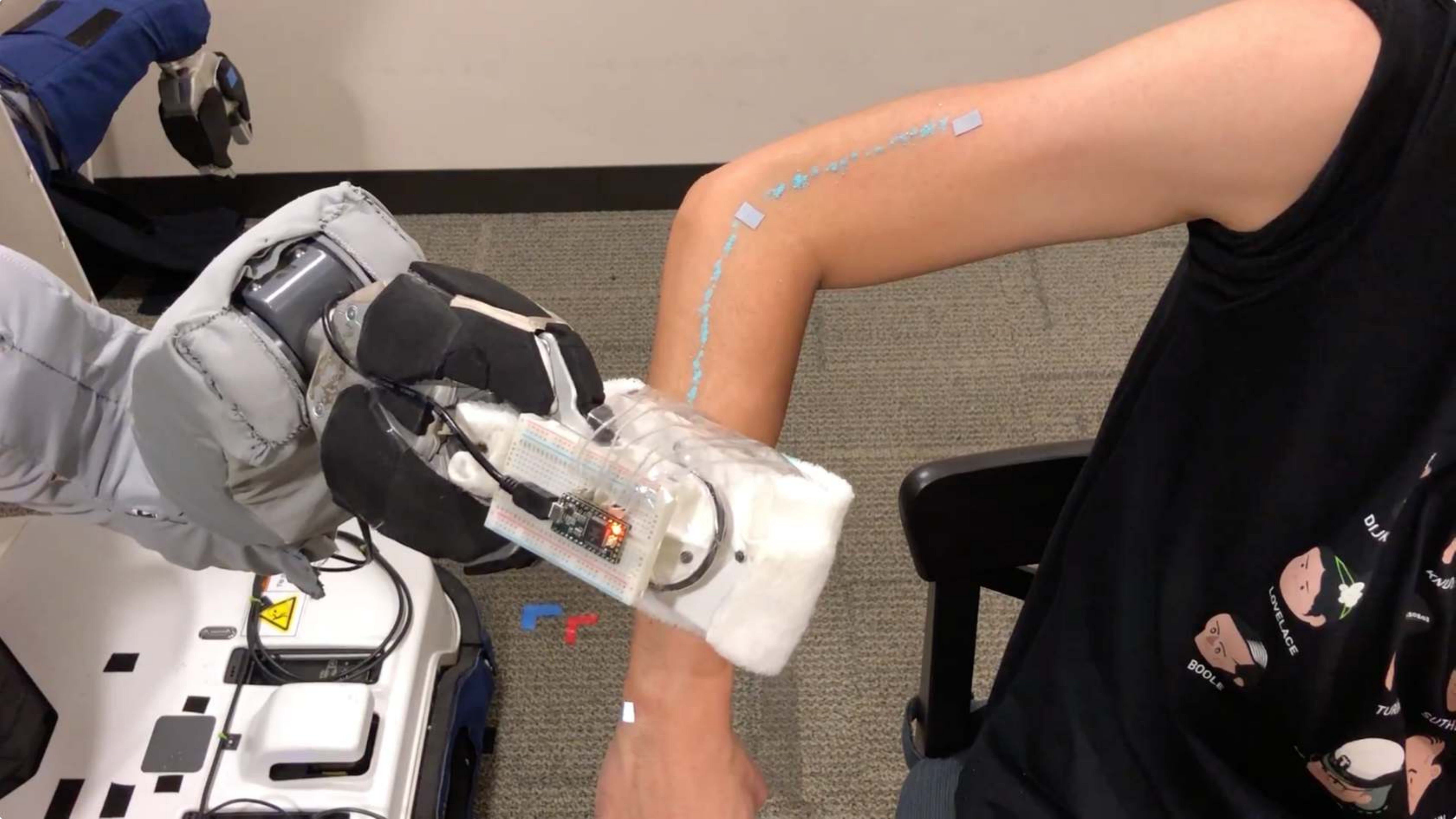}
\includegraphics[width=0.24\textwidth, trim={10cm 0cm 5cm 0cm}, clip]{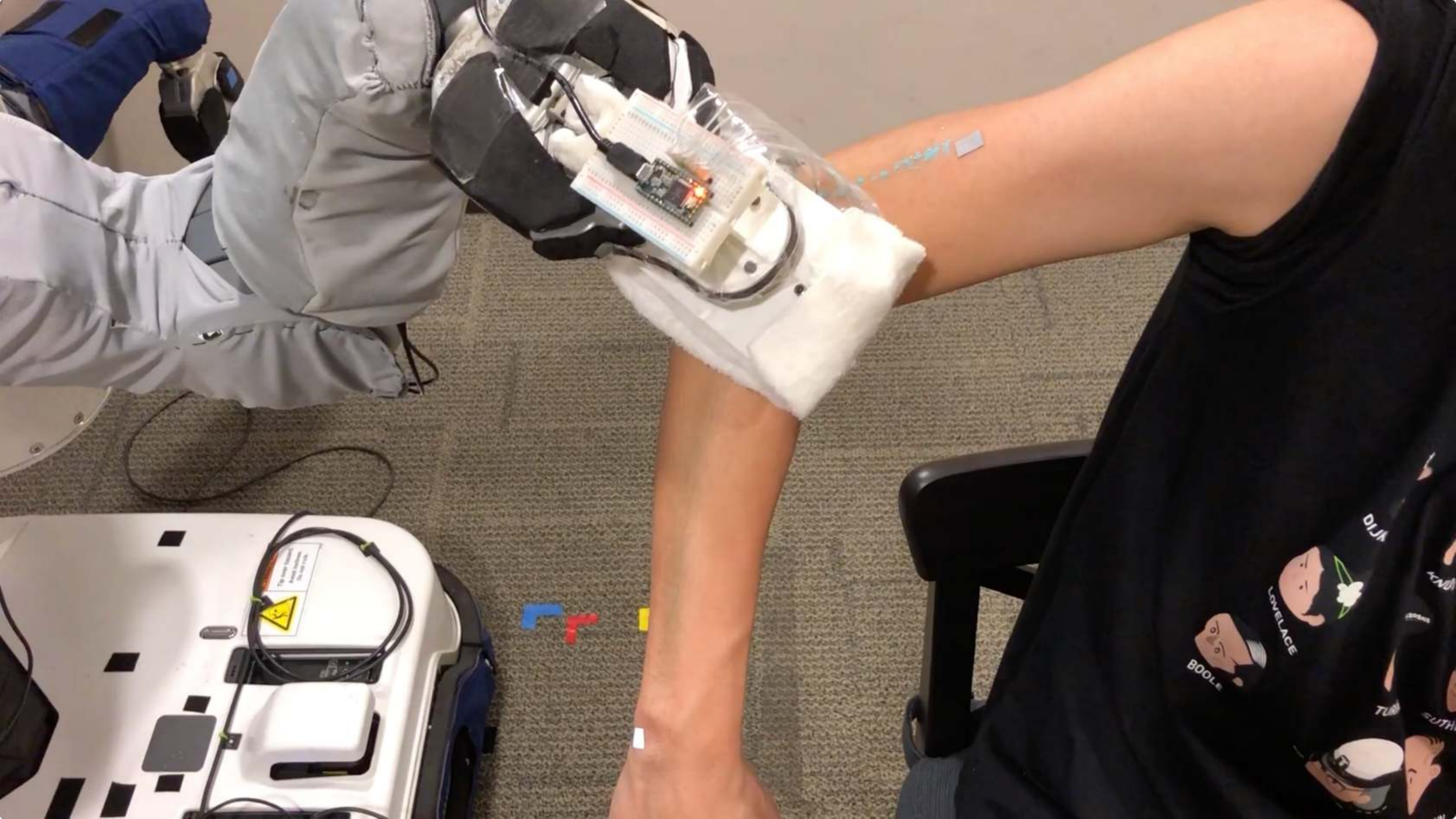}
\includegraphics[width=0.24\textwidth, trim={10cm 0cm 5cm 0cm}, clip]{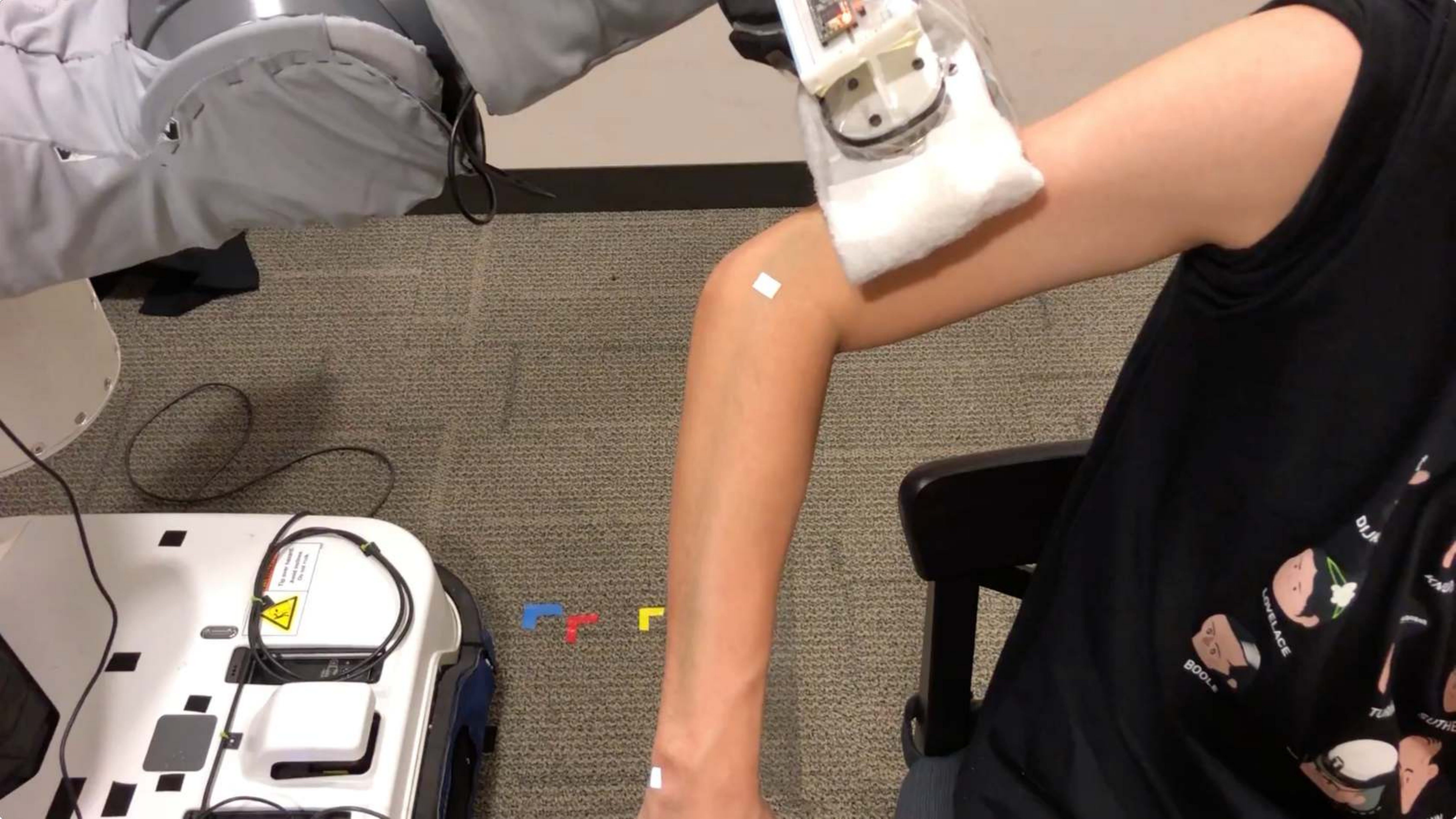}
\vspace{-0.2cm}
\caption{\label{fig:arm_wiping}The PR2 performs a bathing task by using a wet washcloth with capacitive sensing to wipe blue powder off of a participant's arm. The participant's elbow was bent at $\sim$90 degrees and the participant's forearm was tilted 60 degrees downwards towards the ground (see Fig.~\ref{fig:intro_wiping}).}
\vspace{-0.4cm}
\end{figure*}

We then instructed participants to perform lateral arm motions by shifting their arm left to right and bending their elbow, such that their hand remained within 20~cm to the left or right of their starting arm pose. Examples of this vertical arm motion can be found in the supplementary video.
Fig.~\ref{fig:lateral_distance} depicts how the end effector tracked the lateral arm movement for a single participant. On average, the estimated relative lateral position of the end effector, $\hat{p}^{}_y$, remained 2.6~cm from the closest point on the arm throughout the entire dressing trial.
These results remained consistent across all participants, as the robot successfully pulled the sleeve up to each participant's shoulder in all 8 trials.

\begin{figure}
\centering
\includegraphics[width=0.48\textwidth, trim={1cm 19.5cm 2.5cm 3.5cm}, clip]{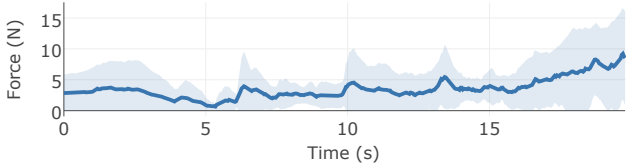}
\vspace{-0.6cm}
\caption{\label{fig:leg_wiping_forces}Average force applied by the PR2's end effector as the robot cleaned participants' legs. Results are averages across all participants with shaded regions representing one standard deviation.}
\end{figure}

\begin{figure}
\centering
\includegraphics[width=0.48\textwidth, trim={1cm 19.5cm 2.5cm 3.5cm}, clip]{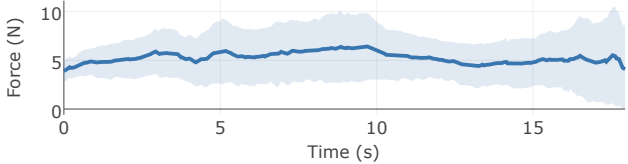}
\vspace{-0.6cm}
\caption{\label{fig:arm_wiping_forces}Average force applied by the PR2's end effector as the robot cleaned participants' forearms and upper arms. Results are averages across all participants with shaded regions representing one standard deviation.}
\vspace{-0.4cm}
\end{figure}

\subsection{Maintaining Continuous Contact}
\label{sec:bathing}

In this section, we describe how the robot can use this same capacitive sensing approach to assist with tasks that require continuous contact with a person's body, such as bathing. Bathing, and the process of cleaning the body with a wet cloth, is fundamental to bed baths, an assistive task common among people who have difficulty leaving bed to shower or bathe.
To assist with bathing, we attached a soft wet washcloth to the bottom of the capacitive sensor, which the robot uses to wipe the surface of a person's body.

Due to the electrical properties of wet cloth, measurements from the capacitive sensor are substantially different from those observed during dressing. As a result, we repeated the data collection and model training procedures, described in Section~\ref{sec:generalized_model}, with a wet washcloth attached to the capacitive sensors. In doing so, we show that our process for training a model on capacitance measurements is both repeatable and can extend to other assistive tasks that require sensing the human body.

We evaluated our capacitive sensing approach with two distinct bathing tasks: cleaning both the arm and the leg of a participant.
To better visualize the robot cleaning a limb, we placed a small amount of vibrant blue powder across the limb, which the robot wiped off during a trial. 
Since the robot makes contact with a participant and the force it applies is distributed across a large soft washcloth, we increased the force threshold in Algorithm~\ref{alg:control} to 20~N.

When assisting with wiping a participant's leg, the robot's end effector began above a participant's thigh, and moved downwards towards the ankle. We asked participants to place their leg on a footstool and to bend their knee at a 30 degree angle, which forced the end effector to rotate around the knee and match the orientation of the lower leg to succeed. 
Fig.~\ref{fig:leg_wiping} portrays a sequence of images from a successful leg wiping trial.
We note that sensing the human body through a wet washcloth poses a challenge for capacitive sensing, yet despite this difficulty, our capacitive sensing approach successfully cleaned all visible blue powder off a participant's leg for 6 of the 8 trials across all 4 participants.

Finally, when the robot assisted with wiping a participant's arm, the robot's end effector began above a participant's hand and moved inwards towards the shoulder. During this scenario, participants held their upper arm parallel to the ground, with a 90 degree elbow bend, and with their forearm and hand tilted 60 degrees downwards towards the ground, as demonstrated in Fig.~\ref{fig:intro_wiping}. An image sequence of the robot wiping a participant's forearm and upper arm can be seen in Fig.~\ref{fig:arm_wiping}. Overall the robot succeeded to clean all visible powder off a participant's arm in 7 of the 8 total trials.

From these image sequences, we can observe that the robot recognized an orientation change at the elbow and rotated to match the estimated orientation of the upper arm.
Fig.~\ref{fig:leg_wiping_forces} and \ref{fig:arm_wiping_forces} present the average force the end effector applied to participants' legs and arms, respectively. We can observe that the washcloth is continually in contact with a participant's limb, while applying less than 6~N of force on average.

\section{Discussion and Conclusion}

We presented a multidimensional capacitive sensor capable of sensing the relative pose of a person's limb. Unlike prior control approaches that rely on vision or force feedback, capacitive sensing enables a robot to sense the human body through some opaque materials and track human motion before the robot applies forces to a person's body.

We detailed an approach to train a data-driven model on capacitance measurements from a single participant, which estimates the relative position of the closest point on the surface of a person's limb, as well as pitch and yaw orientation between a robot's end effector and the central axis of the limb near the point. 

While our model generalized well to assisting multiple people, there are a few limitations to this approach. First, a model trained on a single person may not always generalize to all people, such as those with very small or large limbs. In these cases, it would be beneficial to train on capacitance measurements collected over multiple people with a wide variety of limb sizes and shapes. Further, our capacitive sensors do not have active shielding and thus may be susceptible to electromagnetic interference (EMI). Some conductive objects or clothing may also inhibit capacitance measurements, such as jewelry or metal decorations.


We demonstrated how this capacitive sensing approach enables a PR2 to assist with two activities of daily living that many older adults with physical disabilities require assistance with---dressing and bathing.
Our approach runs in real time, using only the PR2's on-board CPUs, which enabled the robot to assist with dressing by using capacitive sensing to track human limb movement and follow the contours of participants' arms to pull on the sleeve of a hospital gown. In addition, we demonstrated that our data collection and training process is repeatable and can be used to build customized models for new assistive tasks that require sensing the human body, such as bathing. 


Overall, multidimensional capacitive sensing offers a promising approach for robots to sense the human body and track human motion when providing assistance that requires physical human-robot interaction.






\section*{Acknowledgment}

\textit{We thank Ari Kapusta for his assistance with this work. This work was supported by NSF award IIS-1514258, NSF award DGE-1545287, and the NSF GRFP under Grant No. DGE-1148903. 
Dr. Kemp is a cofounder, a board member, an equity holder, and the CTO of Hello Robot, Inc., which is developing products related to this research. This research could affect his personal financial status. The terms of this arrangement have been reviewed and approved by Georgia Tech in accordance with its conflict of interest policies.}

\bibliographystyle{IEEEtran}

\end{document}